\pdfoutput=1

\documentclass[11pt]{article}

\usepackage[]{acl}
\usepackage{float}
\usepackage{times}
\usepackage{latexsym}

\usepackage{tcolorbox}
\usepackage{adjustbox}
\usepackage{booktabs}
\usepackage{graphicx}
\usepackage{array} 
\usepackage{algorithm}
\usepackage{algpseudocode}
\usepackage{multirow}
\usepackage[T1]{fontenc}

\usepackage[utf8]{inputenc}

\usepackage{microtype}

\usepackage{inconsolata}

\usepackage{graphicx}

\usepackage{xcolor}
\definecolor{ed}{RGB}{225,0,100}

\usepackage{titlesec}

\titlespacing\section{0pt}{5pt plus 3pt minus 2pt}{5pt plus 2pt minus 2pt}
\titlespacing\subsection{0pt}{5pt plus 3pt minus 2pt}{5pt plus 2pt minus 2pt}
\titlespacing\subsubsection{0pt}{12pt plus 4pt minus 2pt}{0pt plus 2pt minus 2pt}

\title{Evaluating the Performance of Large Language Models via Debates}

\author{Behrad Moniri \qquad Hamed Hassani\qquad Edgar Dobriban\\
University of Pennsylvania\\
\normalsize\texttt{\{bemoniri, hassani\}@seas.upenn.edu, dobriban@wharton.upenn.edu}
}
\begin{document}
\maketitle
\begin{abstract}
Large Language Models (LLMs) are rapidly evolving and impacting various fields, necessitating the development of effective methods to evaluate and compare their performance. 
Most current approaches for performance evaluation are either based on fixed, domain-specific questions that lack the flexibility required in many real-world applications, or rely on human input, making them unscalable. To address these issues,  we propose an automated benchmarking framework based on debates between LLMs, judged by another LLM. This method assesses not only domain knowledge, but also skills such as argumentative reasoning and inconsistency recognition. We evaluate the performance of various state-of-the-art LLMs using the debate framework and achieve rankings that align closely with popular rankings based on human input, eliminating the need for costly human crowdsourcing.
\end{abstract}

\section{Introduction}

Although still in their infancy, large language models (LLMs) have emerged as a tool with the potential to transform human-computer interaction and significantly impact various aspects of work and daily life (see e.g., \citet{bubeck2023sparks}, etc.). 

Due to this widespread use and the existence of a wide variety of language models, it is crucial to establish a standardized method for evaluating and ranking these models based on their performance. Improved evaluation will provide guidance for future interaction design and implementation.  There are multiple general approaches for evaluating evaluate the performance of LLMs.

The first approach is a \textit{static} approach which involves evaluating models based on a fixed set of pre-determined questions (benchmarks). Many such benchmarks have been proposed to evaluate the performance of LLMs in various domains, such as medical applications (\citet{singhal2023large, cascella2023evaluating, lievin2024can}), legal applications (\citet{hendrycks2021cuad, guha2024legalbench,katz2024gpt}), trustworthiness (\citet{chao2024jailbreakbench, zhang2023safetybench}), reasoning abilities (\citet{sawada2023arb, valmeekam2022large}), and coding abilities (\citet{liu2024your, du2024evaluating, carlinibenchmark}). See \citet{chang2024eval} for a detailed survey. However, these benchmarks are limited, mainly because they may become contaminated over time as new language models are introduced with the benchmarks potentially included as their training data (see e.g., \citet{bubeck2023sparks, ibrahim2024beyond}). Thus, rankings derived using these methods may not generalize to other, new tasks and questions, even within the same domain.

The second approach to LLM evaluation is a \textit{human-based} approach in which human evaluators are asked to interact with and compare models by prompting, then ranking their performance based on their responses. An example is \textit{Chatbot Arena}, recently introduced by \citet{chiang2024chatbot}, which evaluates LLMs using human feedback collected through crowd-sourcing.
Chatbot Arena has attracted significant attention and media coverage (see, e.g., \citet{ChatGPTreigns, measurementissue}). 
Human-based approaches can effectively resolve the problem of data contamination in the static approaches. However, their reliance on human input limits their scalability. Designing suitable prompts and reading (often long) responses from various models can be very expensive and time-consuming.

The third approach to model evaluation is a \textit{game-based} approach which bypasses the need for  human evaluation by designing a structured \textit{game} in which models compete against each other. The game is crafted with automatically checkable  winning criteria, so the winner can be determined automatically without human intervention. The models are then ranked based on their performance in the game. The game should be designed in a way to require the relevant skills expected of language models so that the performance in these games serves as a proxy for the model's overall abilities. As a result, a good choice of the game is very important for the success of game-based approaches. 

\subsection{Evaluation via Debates}

Debate has been a longstanding tradition since antiquity \cite[Historical Supplement]{sep-argument}. 
It serves as a structured forum for testing conversational and reasoning skills,
and has been integral to philosophical discourse, legal proceedings, and civic engagement (\citet{proksch2015politics, holbrook1999political, benoit2003meta}). Although success in debates requires mastery of the debate topic and domain knowledge, it also needs skills such as defining the problem, identifying and challenging assumptions, recognizing inconsistencies, and prioritizing the relevance of various details within the overall argument \cite{roy2005debating, kennedy2009power}. Such skills are crucial for the effective application of LLMs, which motivates using debate settings in game-based approach to LLM evaluation.

In this paper, we take the game-based approach to model evaluation and design the game to be a structured \textit{debate} between competing language models, judged by a pre-specified language model.  In this framework, models debate on a set of predetermined topics. The script of each debate is then given to a pre-determined judge, which is also an LLM, to evaluate and score the arguments presented by each side. Based on these evaluations, the language model with the better overall performance is announced as the winner.  Through experiments with state-of-the-art language models, we demonstrate that this fully automated setup can be used to rank these LLMs, producing rankings generally consistent with those of Chatbot Arena, which is one of the most widely used methods of LLM evaluation.


\begin{figure}[h!]
    \centering
    \caption{A snippet of debates. Two language models engage in debates on a list of topics, and a judge model announces the winner for each topic. The language model with the most wins across all topics is declared the overall winner.}
    \vspace{-0.2cm}
    \label{fig:debate_example}
    \includegraphics[width=1\columnwidth]{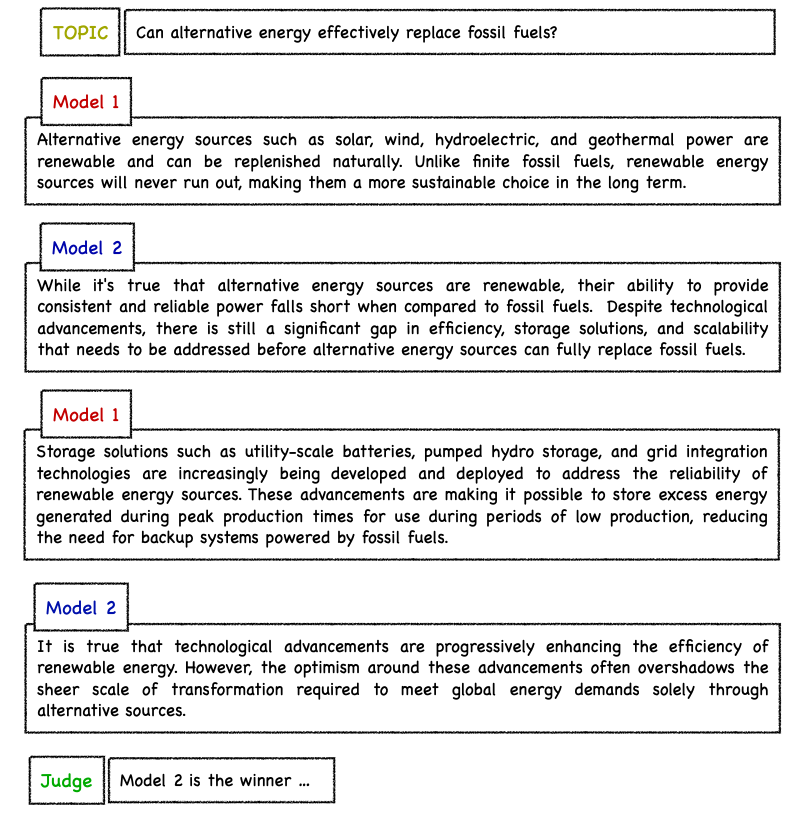}

    \vspace{-0.7cm}
\end{figure}

\section{Related Works}
Collaboration and debates have been used as a framework to enhance the performance of language models. \citet{liang2023encouraging, deRidder} propose  multi-agent debate frameworks in which models express their solutions to a problem to encourage divergent thinking. \citet{xiong2023examining} uses a formal debate framework to examine whether LLMs can collaborate to reach a consensus for a shared goal. \citet{chern2024can} proposes multi-round discussions between models to assist human annotators in finding the most capable LLM to be used as an evaluator. \citet{khan2024debating} shows that having access to the a debate between two strong LLMs that posses the necessary information to answer the question of the topic will result in more truthful answers from a weak LLM or a human expert that do not have that piece of information. 
\citet{debate_education} study the applications of LLMs in debate education. See also \citet{pham2023let,fu2023improving,madaan2024self,chao2024jailbreakbench} for other results leveraging the interactions between language models to achieve a given goal.


\section{Debate Framework}
Assume that  $\mathcal{T}$ is a  pre-defined list of debate topics and $\mathcal{LM}$ is a set of large language models that we want to rank based on their performance. The topics are open-ended questions like \textit{``Can alternative energy effectively replace fossil fuels?"} that have two possible sides.  We assume that we have \textit{black-box query access} to all the language models. The  ranking is based on multiple debates between the language models on different topics. 

Each debate is a multi-round interaction on a topic $t \in \mathcal{T}$ between two models, a language model $\mathrm{LM}_1\in \mathcal{LM}$ that goes first and support one side of the argument, and a language model $\mathrm{LM}_2 \in \mathcal{LM}$ that goes second and supports the other side. In the first round, $\mathrm{LM}_1$ is asked to start the debate by providing the arguments to supports their side based on logic, facts, and evidence. Then, in the next round $\mathrm{LM}_2$ responds, refutes the arguments raised, and provides new evidence supporting the other side. The debate between the models continues for a predefined number $T$ of rounds. Finally, the model $\mathrm{LM}_2$ is asked to conclude the debate. All prompts used in our experiments can be found in Section~\ref{sec:sys_prompts}. At the end of each debate, the script of the $T$ rounds is given to a judge LLM, 
which is asked to consider specific pre-defined factors, 
to score $\mathrm{LM}_1$ and $\mathrm{LM}_2$, and to announce the winner.

In this framework, to compare two models $\mathrm{LM}_a$ and $\mathrm{LM}_b$ from the set $\mathcal{L}$, we conduct two debates on each topic $t \in \mathcal{T}$. In the first debate, we set $\mathrm{LM}_1 = \mathrm{LM}_a$, and in the second debate, we set $\mathrm{LM}_1 = \mathrm{LM}_b$. The judge language models evaluate both debates.
Based on their assessments, one of the models $\mathrm{LM}_1$ or $\mathrm{LM}_2$ is declared the winner for the topic $t$, or the result is a draw. We run two debates because the two sides of the argument might not be equally hard to argue for. Also, the judge could be biased and favor the model that goes first (or last). By running the debate with the role of the models flipped, we account for these biases.

This process is repeated for all topics $t \in \mathcal{T}$. The model with the highest number of wins across different topics is declared the overall winner. Finally, after comparing all pairs $(\mathrm{LM}_a,\mathrm{LM}_b)$ of models, the overall ranking of the models in $\mathcal{L}$ is generated. 

\subsection{Algorithm Details}
Since LLMs are typically not directly trained to debate, specifying the rules of the task is crucial. 
To effectively do this, we provide a detailed system prompt to the models. 
The systems prompts ask the models to support a side;
ask that the arguments and rebuttals should be backed by logic, facts, and evidence; and that the answers should  be convincing, factual and concise. The history of all previous rounds of the debate is given to the debating LLMs.

The same is also true for the judge language model. We set a detailed system prompt for the judge so that it considers factors such as  clarity of arguments, factuality and use of evidence, rebuttal and counterarguments, logical consistency, persuasiveness, conciseness, and coherence in the evaluation. 
Further, in the system prompt for the judge, we specify the exact format for the output. The script of the debate is given to the judge model using its prompt and the judge model announces a winner.

\subsection{The Choice of the Judge LLM}
\label{sec:LLMasJudge}
LLMs have been used as judges for various applications (see e.g., \citet{zheng2024judging,chen2024humans,chao2024jailbreakbench,kim2023prometheus,hada2024large,hada2024metal,wu2023style,kim2024prometheus}).

Because of the symmetry in the game, the evaluation based on the game is \textit{fair} regardless of the strength or weakness of the judge model. However, when we use weaker language models as a judge, we give preference to language models that are stronger at \textit{convincing the judge that their argument is coherent}, instead of the models that are \textit{actually} giving coherent responses. In other words, models that are able to generate responses that are marked by the judge to be more coherent are scored higher. Note that the same is true for human judges. For example, in Chatbot Arena, users  select the response of an LLM not necessarily based on whether it is more ``coherent", but based on whether it is more ``coherent". Although this might not be the right metric for all applications, we argue that for the conversation-based tasks that chatbots are used for everyday, the persuasiveness is the key ability users seek when choosing the language model. For example, see OpenAI o1 System Card \cite[Section 4.7.1]{jaech2024openai}.

 For the task of evaluating debates, \citet{liu2024empirical} showed that of GPT-4 outperforms humans and other state-of-the-art LLMs fine-tuned on extensive datasets in debate evaluation.  More generally, GPT-4 models have consistently been demonstrated to  closely match with human intentions when acting as a judge
 and have been used as a judge extensively in LLM literature (see e.g., \citet{zheng2024judging, achiam2023gpt} and references therein). Based on these findings, we choose GPT-4 as the judge model in our experiments. In Section~\ref{sec:other}, conduct some experiments with Llama-3-70b as the judge and demonstrate that the rankings do not change significantly when Llama-3-70b is used as judge. Also, we conducted  an experiment where a human judge also asked to determine the winner of debates by reading them. We showed that the winners chosen by GPT-4 is  consistent with the winners chosen by the human evaluator.

In this paper, we evaluate the overall conversation ability of language models with general questions and use general-purpose LLMs as judge. However, we note that if all the debate topics are from a specific subject, choosing a Retrieval-Augmented Generation LLM (see e.g., \cite{lewis2020retrieval}) with a proper knowledge-base might be better suited for determining the factuality of the claims by different sides of the debate. 

\section{Experimental Results}
In this section, we consider the list of debate topics from Section~\ref{sec:topics} and run debates with four rounds, i.e., $T=4$. As the debating models, we use {Llama-2-7b}, {Llama-2-13b}, {Llama-2-70b} \cite{touvron2023llama}, {Llama-3-70b} \cite{llama-3}, {Vicuna-7b-v1.5},{Vicuna-13b-v1.5} \cite{vicuna2023}, {Mixtral-8x7B-Instruct-v0.1} \cite{jiang2024mixtral}, {gpt-4-0125-preview}, and {gpt-3.5-turbo-0125} \cite{achiam2023gpt}.  To access GPT models, we use the OpenAI API. For other models, we use the API from {\texttt{https://www.together.ai/}}.

\subsection{Rankings}
\label{sec:rankings}
\begin{figure}[h!]
    \centering
    \vspace{-0.2cm}
    \caption{Overall ranking of LLMs with GPT-4 as judge.}
    \label{fig:ranking}
    \includegraphics[width=0.8\linewidth]{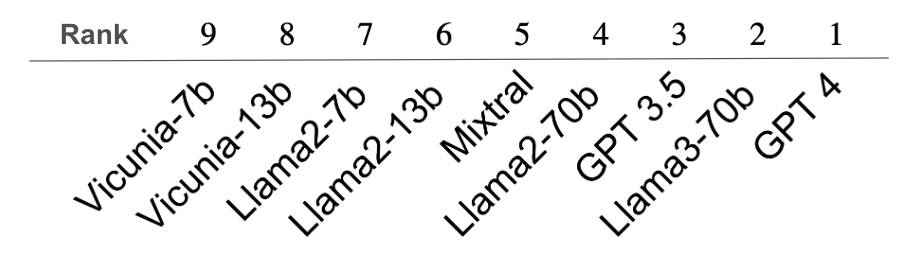}
    \vspace{-0.3cm}
\end{figure}

We run a total of $50$ debates on $25$ topics (Section~\ref{sec:topics}) between each pair of models. In this section we use gpt-4-0125-preview as the judge. On each topic, the model that wins both rounds is considered the winner. Table~\ref{table:experiments-gpt4} (Left) shows the number of wins in the debates between various model pairs. For example, Llama-2-7b has won in no topics against Llama-2-70b, whereas Llama-2-70b has won in nine debates against Llama-2-7b. The detailed results in different topics can be found in Section~\ref{sec:results}. See Table~\ref{table:experiments-gpt4} and Figure~\ref{fig:ranking}

From these results, as a sanity check, it is seen that for different families of models (Llama-2, GPT, Vicuna), models with more parameters rank better; i.e., Llama-2-70b ranks better compared to  Llama-2-7b. Also, newer generations of each model rank better than their older counterpart; i.e., GPT-4 ranks better compared to GPT-3.5. Similarly, Llama-3-70b ranks better compared to Llama-2-70b. 


\begin{table}
\caption{The results from Section~\ref{sec:rankings} with GPT-4 as judge. Numbers indicate the number of topics in which each model has won against other models}
\label{table:experiments-gpt4}
\begin{minipage}{1\columnwidth}
\centering
\resizebox{6.5cm}{!}{
\begin{tabular}{r@{\hspace{0.5cm}}|ccc@{\hspace{0.7cm}}cc@{\hspace{0.7cm}}cc@{\hspace{0.7cm}}c@{\hspace{0.7cm}}cc}
\toprule
 & 
\rotatebox{90}{Llama-2-7b} & 
\rotatebox{90}{Llama-2-13b} & 
\rotatebox{90}{Llama-2-70b} & 
\rotatebox{90}{Llama-3-70b} & 
\rotatebox{90}{Vicuna-7b} & 
\rotatebox{90}{Vicuna-13b} & 
\rotatebox{90}{Mixtral-8x7B} & 
\rotatebox{90}{GPT-3.5} & 
\rotatebox{90}{GPT-4} \\
\midrule\\
Llama-2-7b    &1-0&0-6&0-9&0-11&9-0&6-2  &1-8 &0-5 &0-24  \\[0.5ex]
Llama-2-13b   &   &0-2&0-7&2-13&14-0&8-0 &0-5 &0-7 &0-22  \\[0.5ex]
Llama-2-70b   &   &   &1-1&0-9 &13-0&12-0&2-1 &1-2 &0-21  \\[2ex] 
Llama-3-70b   &   &   &   &0-1 &23-0&17-0&6-0 &5-0 &0-13  \\[2ex] 
Vicuna-7b     &   &   &   &    &2-1&0-3  &0-11&0-18&0-24  \\[0.5ex]
Vicuna-13b    &   &   &   &    &   &0-0  &0-13&0-13&0-23  \\[2ex]
Mixtral-8x7B  &   &   &   &    &   &     &3-1 &0-3 &0-24  \\[2ex]
GPT-3.5       &   &   &   &    &   &     &    &3-2 &1-14  \\[0.5ex]
GPT-4         &   &   &   &    &   &     &    &    &2-2   \\[0.5ex]
\bottomrule
\end{tabular}}
\end{minipage}\hfill 
\end{table}
Comparing this ranking with the rankings available on the Chatbot Arena leaderboard website, accessed on June 14th, 2024, we see that the rankings in Table~\ref{table:experiments-gpt4} are generally consistent with Chatbot Arena. Specifically, the normalized Kendall tau distance between these two rankings is \textbf{0.0833}. This distance takes values in $[0,1]$ where 0 means identical rankings and 1 means reversed rankings.

\subsection{Other Experiments}
\label{sec:other}
\begin{figure}[h!]
\vspace{-0.1cm}
    \centering
    \caption{The percentage of the times each model won against another model for each of the six reasons.}
    \label{fig:histogram}
    \includegraphics[width=\linewidth]{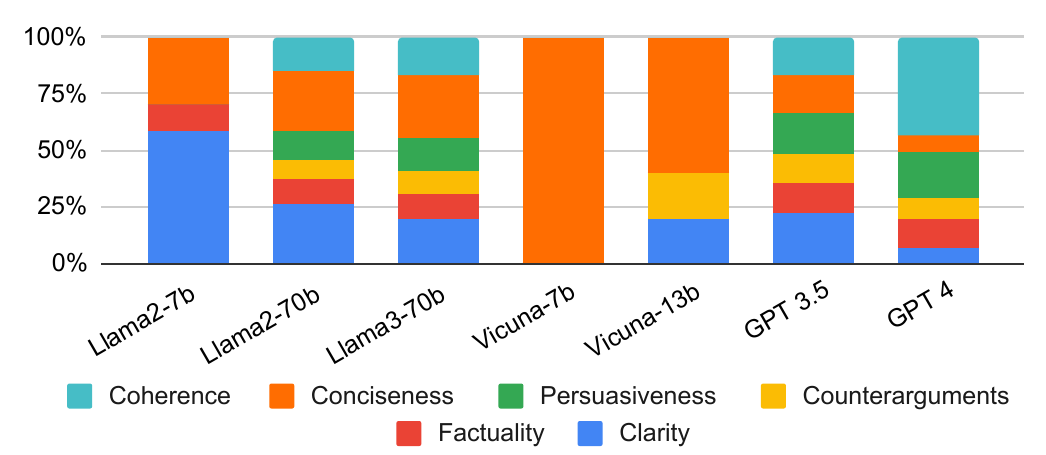}
    \vspace{-0.3cm}
\end{figure}

\paragraph{Analysis of Content.}
\label{sec:analysis_debate}
We prompt the judge (GPT-4) model to state the main reason for its choice among "clarity, factuality, counterarguments, persuasiveness, conciseness, and coherence."  Figure~\ref{fig:histogram} illustrates the stated reasons for which different models won the debates against their opponents. This shows that clarity and coherence have been the most decisive factors in the decisions. It also reveals that, for example, the responses by GPT-4 were seen by the judge as being more coherent, whereas the arguments by Vicuna-7b were seen to be more concise.
\paragraph{Human as Judge.} In this experiment, we asked three human student volunteers to read the contents of the debates between Llama-2-13b and Llama-2-70b and judge the debates. The human evaluators each judged a total of 50 debates and their judgments matched the results of Section C.11 in 81.3\% of the time. In particular, each participant agreed with the judgment from Section C.11 in 43, 41, and 38 debates out of 50.

\paragraph{Llama-3 as Judge.} Finally, we conduct similar experiments, but with Llama-3-70b as the judge. For demonstration, we only repeat the debates between Llama-2-13b with all other models. The score of Llama-2-13b against other models with Llama-3-70b as judge is shown in Table~\ref{tab:other-judge}. The winners in this experiment are identical to the winners announced by GPT-4, in all but one opponent models (Mixtral).
\begin{table}[h!]
\centering
\vspace{-0.2cm}

\caption{Score of Llama-2-13b vs. other models (first numbers for Llama-2-13b), with Llama-3-70b as judge.}
\vspace{-0.2cm}
\label{tab:other-judge}
\resizebox{5cm}{!}{
\begin{tabular}{cc|cc}
\textbf{Model} &  \textbf{Score} & \textbf{Model} & \textbf{Score} \\
\midrule
Llama-2-7b     & 7-0  & Vicuna-13b & 17-0\\
Llama-2-70b    & 0-4  & Mixtral-8x7B & 8-4\\
Llama-3-70b    & 0-6  & GPT-3.5 & 0-4\\
Vicuna-7b      & 16-2  & GPT-4 & 0-21
\end{tabular}}
\vspace{-0.3cm}
\end{table}


\section{Conclusion}
In this paper, we developed an automated framework to rank the performance of LLMs, 
based on a multi-round debate between LLMs on different topics, and an evaluation by a judge LLM. 
We showed that this framework can yield rankings consistent with rankings that rely on human crowdsourcing, while being more scalable.

\section{Limitations}
\vspace{0.2cm}
\paragraph{List of Topics.} The method proposed in this paper requires human input to create the list of debate topics. While this is significantly less time-consuming and less expensive than reading and scoring the debates, it can still pose some scalability issues. We leave the task of automating topic generation for future work.

\paragraph{LLM as Judge.}
The debate framework introduced in this paper is a game-based approach to model evaluation and uses a judge LLM as a part of the game. Because of the symmetry in the game design, the judge cannot bias the game; however, if a weak LLM is chosen as the judge, it is possible that the judge may not be capable of fully evaluating qualities such as the ``factuality" of the arguments. In such cases, the winner is determined by the abilities of different models to \textit{convince} the judge that they are more factual, instead of \textit{actually} giving more factual answers. Although this still demonstrates the abilities of that language model, it might not necessarily be the qualities that we try to evaluate. Note that the same is true for human judges. For example, in Chatbot Arena, users  select an LLM not necessarily based on whether it is generated more ``factual" responses, but based on whether it can generate more ``convincing" outputs.

In our experiments in Section~\ref{sec:other}, it was shown that the decisions are mostly based on factors such as coherence, conciseness, or clarity that are easier for a language model to evaluate. However, despite the evidence (see e.g., \citet{liu2024empirical}) that models such as GPT-4 perform well on debate evaluation tasks, in general this limits the applicability of the debate framework to evaluate the ability of models in tasks other than the general conversational ability that was studied here, where factors such as factuality are more important. 

\paragraph{Choice of Language.}
Our method has been primarily evaluated on English, which is a language with relatively limited morphological complexity. This could restrict the applicability of our approach to other languages.


 \bibliography{custom}
\newpage

\onecolumn
\appendix
\section{System Prompts and Prompts}
\label{sec:sys_prompts}
In this section, we list the system prompts and prompts used in the experiments.
\subsection{Debating Models}
The first language model $\mathrm{LM}_1$ is asked to start the debate. This language model will have the following system prompt.
\begin{tcolorbox}\underline{\textbf{System Prompt for $\mathrm{LM}_1$}}\vspace{0.2cm}

We are having a debate and the topic is ``\textbf{TOPIC}". You are representing ``\textbf{SIDE 1}" and you should zealously support it. This the first round and you are going first. You must bring-up arguments supporting your side backed by logic, facts and evidence. Your answer should also be convincing, factual and concise.
\end{tcolorbox}

We will prompt $\mathrm{LM}_1$ as follows.
\begin{tcolorbox}\underline{\textbf{Prompt for $\mathrm{LM}_1$}}\vspace{0.2cm}

Please start the debate.
\end{tcolorbox}

In the second part of the debate, the second language model $\mathrm{LM}_2$  will be given the response of $\mathrm{LM}_1$. The second model is asked to provide a rebuttal to the points raised by $\mathrm{LM}_1$. Also, it is asked to provide new arguments supporting the second side. This language model will have the following system prompt.

\begin{tcolorbox}\underline{\textbf{System Prompt for $\mathrm{LM}_2$}}\vspace{0.2cm}

We are having a debate and the topic is ``\textbf{TOPIC}". You are representing ``\textbf{SIDE 2}" and you should zealously support it. The other side has started the debate and you will be given their arguments. You must first provide a rebuttal to the points raised by them, and then provide new arguments supporting your side. All your arguments should be backed by logic, facts and evidence. Your answer should also be convincing, factual and concise.
\end{tcolorbox}
We prompt $\mathrm{LM}_2$ as follows.
\begin{tcolorbox}\underline{\textbf{Prompt for $\mathrm{LM}_2$}}\vspace{0.2cm}

The other side said: ``\textbf{Response of $\mathrm{LM}_1$ in Part 1}".
\end{tcolorbox}

In all following steps, we use the following system prompts for the models.
\begin{tcolorbox}\underline{\textbf{System Prompt}}\vspace{0.2cm}

We are having a debate and the topic is ``\textbf{TOPIC}". You are representing ``\textbf{SIDE}" and you should zealously support it. You will be given all the arguments so far. In your response, you should support your side and refute the points raised by the other side. Your arguments should be backed by logic, facts and evidence. Your answer should also be convincing, factual and concise.
\end{tcolorbox}
We prompt the models $\mathrm{LM}_1$ and $\mathrm{LM}_2$  as follows.

\begin{tcolorbox}\underline{\textbf{Prompt for $\mathrm{LM}_1$}}\vspace{0.2cm}

You initially said: ``\textbf{Response of $\mathrm{LM}_1$ in Part 1}". You the other side responded: ``\textbf{Response of $\mathrm{LM}_2$ in Part 2}". Then you said: ``\textbf{Response of $\mathrm{LM}_1$ in Part 3}", \dots .
\end{tcolorbox}

\begin{tcolorbox}\underline{\textbf{Prompt for $\mathrm{LM}_2$}}\vspace{0.2cm}

The other side initially said: ``\textbf{Response of $\mathrm{LM}_1$ in Part 1}". You then responded: ``\textbf{Response of $\mathrm{LM}_2$ in Part 2}". The other side said: ``\textbf{Response of $\mathrm{LM}_1$ in Part 3}", \dots .
\end{tcolorbox}
\subsection{Judge Model}

A judge language model gives scores from 1 to 10 to each debate participant based on the following criteria: (1) Clarity of arguments (2) Factuality and use of evidence (3) Rebuttal and counterarguments (4) Logical consistency (5) Persuasiveness and impact (6) Conciseness (7) Coherence.  The judges will have the following system prompt.

\begin{tcolorbox}\underline{\textbf{System Prompt for the Judge}}\vspace{0.2cm}

We had a  debate and the topic was ``\textbf{TOPIC}". The two sides in the debate each provided arguments to prove their side and refute the points raised by the opponent. You are a judge for this debate. You should be impartial and as objective as possible. The debate script will be given. You should give a score from 1 to 10 to each side of the debate. In your judgement, you should take into account the following criteria: clarity of arguments, factuality and use of evidence, rebuttal and counterarguments, logical consistency, persuasiveness and impact, conciseness, coherence. Also, you should choose the side who you think is the overall winner. Your answer MUST follow the following format: "side1: [[score of side 1]], side2: [[score of side 2]], winner: [[name of winner]]" where score of side 1 and score of side 2 are numbers from 1 to 10 and name of winner is either "1" or "2".
\end{tcolorbox}

The judge generates the scores given the whole script of the debate, using the following prompt.
\begin{tcolorbox}\underline{\textbf{Prompt for the Judge}}\vspace{0.2cm}

The script of the debate is as follows:
Side 1: ``\textbf{Response of $\mathrm{LM}_1$ in Part~1}".
Side 2: ``\textbf{Response of $\mathrm{LM}_2$ in Part~2}".
Side 1: ``\textbf{Response of $\mathrm{LM}_1$ in Part~3}".
Side 2: ``\textbf{Response of $\mathrm{LM}_2$ in Part~4}".
\end{tcolorbox}

\section{Topic of Debates}
\label{sec:topics}
The  twenty-five debate topics used in the experiments are as follows.
\begin{enumerate}
\setlength\itemsep{0.07em}
\item Can alternative energy effectively replace fossil fuels?
\item Should K-12 students dissect animals in science classrooms?
\item Is artificial intelligence good for society?
\item Should bottled water be banned?
\item Is a college education worth it?
\item Should the United States keep daylight saving time?
\item Should school dress codes be implemented and enforced?
\item Should the drinking age be lowered from 21 to a younger age?
\item Should the election day be made a national holiday?
\item Should the governments use Large Language Models for advice?
\item Should employers be able to mandate vaccinations?
\item Should fighting be allowed in hockey?
\item Should fur clothing be banned?
\item Should genetically modified organisms (GMOs) be grown?
\item Is golf a sport and are golfers athletes?
\item Is homework beneficial?
\item Is the internet “making us stupid?”
\item Should medical aid in dying be legal?
\item Is obesity a disease?
\item Should the penny stay in circulation?
\item Are the Olympic games an overall benefit for their host countries and cities?
\item Is there really a Santa Claus?
\item Should Halloween be moved permanently to Saturday?
\item Should students have to wear school uniforms?
\item Is social media good for society?
\end{enumerate}

\section{Experimental Results}
\label{sec:results}

In this section, we report the experimental results with GPT-4 as judge. Each table is the result of the debates between two models on all topics. \textit{Home} and \textit{Away} correspond to to runs of the debate on each topic, each time with one LLM going first.

\subsection{Llama-2-7b vs Llama-2-7b}
\begin{table}[h!]
\centering
{\scriptsize
\begin{tabular}{c|ccc|ccc}
& \multicolumn{3}{c}{Home}& \multicolumn{3}{c}{Away} \\[0.2cm]
\midrule       
& &Side 1& Side 2& &Side 1 &Side 2   \\ 
Topic  & Winner &Llama-2-7b  & Llama-2-7b  & Winner&Llama-2-7b  & Llama-2-7b   \\ \midrule 
 1& Side 1 &8& 7& Side 1&9& 7 \\
 2& Side 2 &7& 8& Side 2&8& 9 \\
 3& Side 1 &8& 7& Side 1&8& 7 \\
 4& Side 1 &8& 7& Side 1&8& 7 \\
 5& Side 1 &8& 7& Side 2&8& 9 \\
 6& Side 1 &8& 7& Side 1&8& 7 \\
 7& Side 2 &7& 8& Side 2&7& 8 \\
 8& Side 2 &7& 8& Side 2&6& 8 \\
 9& Side 1 &8& 7& Side 1&8& 7 \\
 10& Side 1 &8& 7& Side 1&8& 7 \\
 11& Side 1 &9& 7& Side 1&8& 7 \\
 12& Side 2 &7& 8& Side 2&7& 8 \\
 13& Side 1 &8& 7& Side 1&8& 7 \\
 14& Side 1 &8& 7& Side 1&8& 7 \\
 15& Side 1 &8& 6& Side 1&8& 5 \\
 16& Side 2 &7& 8& Side 2&7& 8 \\
 17& Side 2 &7& 8& Side 2&7& 9 \\
 18& Side 1 &8& 7& Side 1&8& 7 \\
 19& Side 1 &9& 7& Side 1&9& 7 \\
 20& Side 2 &7& 8& Side 2&7& 8 \\
 21& Side 2 &7& 8& Side 2&8& 9 \\
 22& Side 2 &6& 8& Side 2&6& 8 \\
 23& Side 1 &8& 7& Side 1&8& 7 \\
 24& Side 2 &7& 8& Side 2&7& 8 \\
 25& Side 1&8& 7& Side 1&8& 7 
\end{tabular}
}
\end{table}

\newpage
\subsection{Llama-2-7b vs Llama-2-13b}
\begin{table}[h!]
\centering
{\scriptsize
\begin{tabular}{c|ccc|ccc|c}
& \multicolumn{3}{c}{Home}& \multicolumn{3}{c}{Away} & \\[0.2cm]
\midrule       
& &Side 1& Side 2& &Side 1 &Side 2   & \\ 
Topic  & Winner &Llama-2-13b  & Llama-2-7b  & Winner&Llama-2-7b  & Llama-2-13b   & Overall\\ \midrule 
 1 &Llama-2-13b&8& 7 & Llama-2-7b& 8 &7  & Tie\\
2 &Llama-2-7b& 7& 8 & Llama-2-13b& 7 &8  & Tie\\
3 &Llama-2-13b& 8& 7 & Llama-2-7b& 8 &7  & Tie\\
4 &Llama-2-13b& 8& 7 & Llama-2-7b& 8 &7  & Tie\\
5 &Llama-2-13b& 8& 7 & Llama-2-7b& 8 &7  & Tie\\
6 &Llama-2-13b& 8& 7 & Tie & 8 &8  & Llama-2-13b\\
7 &Llama-2-7b& 8& 9 & Llama-2-13b& 7 &9  & Tie\\
8 &Llama-2-7b& 6& 8 & Llama-2-13b& 7 &8  & Tie\\
9 &Llama-2-13b& 8& 7 & Llama-2-7b& 8 &7  & Tie\\
10 &Llama-2-13b& 8& 7 & Llama-2-7b& 8 &7  & Tie\\
11 &Llama-2-13b& 8& 7 & Llama-2-7b& 8 &7  & Tie\\
12 &Llama-2-7b& 7& 8 & Llama-2-13b& 7 &8  & Tie\\
13 &Llama-2-13b& 8& 7 & Llama-2-7b& 8 &7  & Tie\\
14 &Llama-2-13b& 8& 7 & Llama-2-13b& 7&8 & Llama-2-13b\\
15 &Llama-2-13b& 8& 7 & Llama-2-7b& 8 &7  & Tie\\
16 &Llama-2-13b& 8& 7 & Llama-2-7b& 8 &7  & Tie\\
17 &Llama-2-7b& 6& 8 & Llama-2-13b& 7 &8  & Tie\\
18 &Llama-2-13b& 8& 7 & Llama-2-7b& 8 &7  & Tie\\
19 &Llama-2-13b& 9& 7 & Llama-2-7b& 8 &7  & Tie\\
20 &Llama-2-7b& 6& 8 & Llama-2-13b& 7 &8  & Tie\\
21 &Llama-2-13b& 8& 7 & Llama-2-13b& 8 &9  & Llama-2-13b\\
22 &Llama-2-7b& 6& 8 & Llama-2-13b& 6 &8  & Tie\\
23 &Llama-2-13b& 8& 7 & Llama-2-13b& 7 &8  & Llama-2-13b\\
24 &Llama-2-13b& 8& 7 & Llama-2-13b& 8 &9  & Llama-2-13b\\
25 &Llama-2-13b& 8& 7 & Llama-2-13b& 8 &9  & Llama-2-13b\\
\end{tabular}
}
\end{table}

\subsection{Llama-2-7b vs. Llama-2-70b}

\begin{table}[h!]
\centering
{\scriptsize
\begin{tabular}{c|ccc|ccc|c}
&\multicolumn{3}{c}{Home}&\multicolumn{3}{c}{Away} & \\[0.2cm]
\midrule       
&  & Side 1& Side 2&  &Side 1 &Side 2   & \\ 
Topic  & Winner &Llama-2-7b& Llama-2-70b& Winner&Llama-2-70b& Llama-2-7b & Overall\\ \midrule 
1  &Llama-2-7b& 8& 7& Llama-2-70b& 9&7 & Tie\\
2  &Llama-2-70b& 7& 9& Llama-2-7b & 7&8 & Tie\\
3  &Llama-2-7b& 8& 7& Llama-2-70b& 8&7 & Tie\\
4  &Llama-2-7b& 8& 7& Llama-2-70b& 8&7 & Tie\\
5  &Llama-2-70b& 8& 9& Llama-2-70b& 8&7 & Llama-2-70b\\
6  &Llama-2-70b& 7& 8& Llama-2-70b& 8&7 & Llama-2-70b\\
7  &Llama-2-70b& 7& 8& Llama-2-70b& 8&7 & Llama-2-70b\\
8  &Llama-2-70b& 6& 8& Llama-2-7b & 7&8 & Tie\\
9  &Llama-2-7b& 8& 7& Llama-2-70b& 8&7 & Tie\\
10  &Llama-2-70b& 8& 8& Llama-2-70b& 8&8 & Llama-2-70b\\
11  &Llama-2-7b& 8& 7& Llama-2-70b& 8&7 & Tie\\
12  &Llama-2-70b& 7& 9& Llama-2-7b & 8&9 & Tie\\
13  &Llama-2-70b& 7& 8& Llama-2-70b& 8&7 & Llama-2-70b\\
14  &Llama-2-7b& 8& 7& Llama-2-70b& 8&7 & Tie\\
15  &Llama-2-7b& 8& 7& Llama-2-70b& 8&7 & Tie\\
16  &Llama-2-7b& 8& 8& Llama-2-70b& 8&7 & Tie\\
17  &Llama-2-70b& 7& 8& Llama-2-7b & 7&8 & Tie\\
18  &Llama-2-7b& 8& 7& Llama-2-70b& 8&7 & Tie\\
19  &Llama-2-7b& 9& 8& Llama-2-70b& 8&7 & Tie\\
20  &Llama-2-70b& 6& 8& Llama-2-7b & 7&8 & Tie\\
21  &Llama-2-70b& 8& 9& Llama-2-70b& 8&7 & Llama-2-70b\\
22  &Llama-2-70b& 6& 8& Llama-2-70b& 8&7 & Llama-2-70b\\
23  &Llama-2-7b& 8& 7& Llama-2-70b& 8&7 & Tie\\
24  &Llama-2-70b& 7& 8& Llama-2-70b& 8&7 & Llama-2-70b\\
25  &Llama-2-70b& 7& 8& Llama-2-70b& 9&8 & Llama-2-70b\\
\end{tabular}
}
\end{table}

\newpage
\subsection{Llama-2-7b vs. Llama-3-70b}

\begin{table}[h!]
\centering
{\scriptsize
\begin{tabular}{c|ccc|ccc|c}
&\multicolumn{3}{c}{Home}&\multicolumn{3}{c}{Away} & \\[0.2cm]
\midrule       
&  & Side 1& Side 2&  &Side 1 &Side 2   &\\ 
Topic  & Winner &Llama-3-70b& Llama-2-7b& Winner&Llama-2-7b& Llama-3-70b&Overall\\ \midrule 
1  &Llama-3-70b& 9& 8& Llama-2-7b& 9&8&Tie\\
2  &Llama-3-70b& 8& 7& Llama-3-70b& 7&8&Llama-3-70b\\
3  &Llama-2-7b& 8& 9& Llama-3-70b& 8&9&Tie\\
4  &Llama-3-70b& 8& 6& Llama-2-7b& 8&7&Tie\\
5  &Llama-3-70b& 8& 7& Llama-2-7b& 8&7&Tie\\
6  &Llama-3-70b& 8& 7& Llama-3-70b& 7&8&Llama-3-70b\\
7  &Llama-3-70b& 8& 7& Llama-3-70b& 7&9&Llama-3-70b\\
8  &Llama-3-70b& 8& 7& Llama-3-70b& 6&8&Llama-3-70b\\
9  &Llama-3-70b& 8& 7& Llama-2-7b& 8&7&Tie\\
10  &Llama-3-70b& 8& 7& Llama-3-70b& 8&9&Llama-3-70b\\
11  &Llama-3-70b& 8& 7& Llama-2-7b& 8&7&Tie\\
12  &Llama-3-70b& 8& 7& Llama-3-70b& 7&8&Llama-3-70b\\
13  &Llama-3-70b& 8& 6& Llama-2-7b& 8&7&Tie\\
14  &Llama-3-70b& 8& 7& Llama-2-7b& 8&7&Tie\\
15  &Llama-3-70b& 9& 7& Llama-2-7b& 8&7&Tie\\
16  &tie& 8& 8& Llama-3-70b& 8&9&Llama-3-70b\\
17  &Llama-2-7b& 7& 8& Llama-3-70b& 7&9&Tie\\
18  &Llama-3-70b& 8& 7& Llama-2-7b& 8&7&Tie\\
19  &Llama-3-70b& 9& 8& Llama-2-7b& 9&7&Tie\\
20  &Llama-2-7b& 7& 8& Llama-3-70b& 6&8&Tie\\
21  &Llama-3-70b& 8& 7& Llama-3-70b& 7&8&Llama-3-70b\\
22  &Llama-2-7b& 7& 8& Llama-3-70b& 6&8&Tie\\
23  &Llama-3-70b& 8& 7& Llama-3-70b& 7&8&Llama-3-70b\\
24  &Llama-3-70b& 8& 7& Llama-3-70b& 7&8&Llama-3-70b\\
25  &Llama-3-70b& 8& 7& Llama-3-70b& 8&9&Llama-3-70b\\
\end{tabular}
}
\end{table}

\subsection{Llama-2-7b vs. Vicuna-7b-v1.5}

\begin{table}[h!]
\centering
{\scriptsize
\begin{tabular}{c|ccc|ccc|c}
&\multicolumn{3}{c}{Home}&\multicolumn{3}{c}{Away} & \\[0.2cm]
\midrule       
&  & Side 1& Side 2&  &Side 1 &Side 2   & \\ 
Topic  & Winner &Llama-2-7b& Vicuna-7b-v1.5& Winner&Vicuna-7b-v1.5& Llama-2-7b & Overall\\ \midrule 
1  &Llama-2-7b& 9& 7& Vicuna-7b-v1.5& 8&7 & Tie\\
2  &Vicuna-7b-v1.5& 6& 8& Llama-2-7b & 7&8 & Tie\\
3  &Llama-2-7b& 8& 7& tie& 8&8 & Llama-2-7b\\
4  &Llama-2-7b& 8& 7& Vicuna-7b-v1.5& 8&7 & Tie\\
5  &Llama-2-7b& 9& 8& Llama-2-7b & 8&9 & Llama-2-7b\\
6  &Llama-2-7b& 8& 7& Llama-2-7b & 7&8 & Llama-2-7b\\
7  &Vicuna-7b-v1.5& 7& 8& Llama-2-7b & 7&8 & Tie\\
8  &Vicuna-7b-v1.5& 7& 8& Llama-2-7b & 7&8 & Tie\\
9  &Llama-2-7b& 8& 7& Vicuna-7b-v1.5& 8&7 & Tie\\
10  &Llama-2-7b& 8& 7& Llama-2-7b & 7&8 & Llama-2-7b\\
11  &Llama-2-7b& 8& 7& tie& 8&8 & Llama-2-7b\\
12  &Vicuna-7b-v1.5& 7& 8& Llama-2-7b & 6&8 & Tie\\
13  &Llama-2-7b& 9& 7& Llama-2-7b & 7&8 & Llama-2-7b\\
14  &Llama-2-7b& 8& 7& Vicuna-7b-v1.5& 8&7 & Tie\\
15  &Llama-2-7b& 8& 7& Vicuna-7b-v1.5& 8&7 & Tie\\
16  &Llama-2-7b& 7& 6& Llama-2-7b & 7&8 & Llama-2-7b\\
17  &Vicuna-7b-v1.5& 6& 8& Llama-2-7b & 6&8 & Tie\\
18  &Llama-2-7b& 9& 7& Vicuna-7b-v1.5& 8&7 & Tie\\
19  &Llama-2-7b& 9& 7& Llama-2-7b & 8&9 & Llama-2-7b \\
20  &Llama-2-7b& 8& 7& Llama-2-7b & 6&8 & Llama-2-7b \\
21  &Llama-2-7b& 8& 7& Llama-2-7b & 8&9 & Llama-2-7b \\
22  &Vicuna-7b-v1.5& 6& 8& Llama-2-7b & 5&7 & Tie\\
23  &Vicuna-7b-v1.5& 6& 8& Llama-2-7b & 6&8 & Tie\\
24  &Llama-2-7b& 8& 7& Llama-2-7b & 7&8 & Llama-2-7b \\
25  &Vicuna-7b-v1.5& 7& 8& Llama-2-7b & 7&8 & Tie\\
\end{tabular}
}
\end{table}

\newpage
\subsection{Llama-2-7b vs. Vicuna-13b-v1.5}

\begin{table}[h!]
\centering
{\scriptsize
\begin{tabular}{c|ccc|ccc|c}
&\multicolumn{3}{c}{Home}&\multicolumn{3}{c}{Away} & \\[0.2cm]
\midrule       
&  & Side 1& Side 2&  &Side 1 &Side 2   & \\ 
Topic  & Winner &Llama-2-7b& Vicuna-13b-v1.5& Winner&Vicuna-13b-v1.5& Llama-2-7b & Overall\\ \midrule 
1  &Llama-2-7b& 8& 7& Vicuna-13b-v1.5& 8&7 & Tie\\
2  &Vicuna-13b-v1.5& 7& 8& Llama-2-7b & 7&8 & Tie\\
3  &Llama-2-7b& 8& 7& Vicuna-13b-v1.5& 8&7 & Tie\\
4  &Llama-2-7b& 8& 7& Llama-2-7b & 7&8 & Llama-2-7b \\
5  &Llama-2-7b& 8& 7& Vicuna-13b-v1.5& 8&7 & Tie\\
6  &Llama-2-7b& 8& 7& Llama-2-7b & 7&8 & Llama-2-7b \\
7  &Vicuna-13b-v1.5& 8& 9& Llama-2-7b & 7&8 & Tie\\
8  &Vicuna-13b-v1.5& 6& 8& Llama-2-7b & 7&8 & Tie\\
9  &Llama-2-7b& 8& 7& Vicuna-13b-v1.5& 7&6 & Tie\\
10  &Llama-2-7b& 8& 7& Llama-2-7b & 8&9 & Llama-2-7b \\
11  &Llama-2-7b& 8& 7& Vicuna-13b-v1.5& 8&7 & Tie\\
12  &Vicuna-13b-v1.5& 8& 9& Llama-2-7b & 6&9 & Tie\\
13  &Llama-2-7b& 8& 7& Llama-2-7b & 7&8 & Llama-2-7b \\
14  &Llama-2-7b& 8& 7& Vicuna-13b-v1.5& 8&7 & Tie\\
15  &Llama-2-7b& 8& 6& Llama-2-7b & 7&8 & Llama-2-7b \\
16  &Vicuna-13b-v1.5& 8& 8& Llama-2-7b & 7&8 & Tie\\
17  &Vicuna-13b-v1.5& 7& 8& Llama-2-7b & 7&9 & Tie\\
18  &Llama-2-7b& 8& 7& Vicuna-13b-v1.5& 8&7 & Tie\\
19  &Llama-2-7b& 9& 7& Llama-2-7b & 8&9 & Llama-2-7b \\
20  &Vicuna-13b-v1.5& 7& 8& Llama-2-7b & 6&8 & Tie\\
21  &Vicuna-13b-v1.5& 8& 9& Llama-2-7b & 7&8 & Tie\\
22  &Vicuna-13b-v1.5& 7& 8& Llama-2-7b & 7&8 & Tie\\
23  &Vicuna-13b-v1.5& 7& 8& Vicuna-13b-v1.5& 8&7 & Vicuna-13b-v1.5\\
24  &Llama-2-7b& 8& 7& Llama-2-7b & 6&8 & Llama-2-7b \\
25  &tie& 8& 8& Vicuna-13b-v1.5& 8&7 & Vicuna-13b-v1.5\\
\end{tabular}
}
\end{table}

\subsection{Llama-2-7b vs. Mixtral-8x7B}
\begin{table}[h!]
\centering
{\scriptsize
\begin{tabular}{c|ccc|ccc|c}
&\multicolumn{3}{c}{Home}&\multicolumn{3}{c}{Away} & \\[0.2cm]
\midrule       
&  & Side 1& Side 2&  &Side 1 &Side 2   & \\ 
Topic  & Winner &Llama-2-7b  & Mixtral-8x7B  & Winner&Mixtral-8x7B  & Llama-2-7b   & Overall\\ \midrule 
1  &Llama-2-7b& 8 & 7  & Mixtral-8x7B& 8 &7  & Tie\\
2  &Mixtral-8x7B& 7 & 8  & Llama-2-7b& 7 &8  & Tie\\
3  &Llama-2-7b& 8 & 7  & Mixtral-8x7B& 8 &7  & Tie\\
4  &Llama-2-7b& 8 & 7  & Mixtral-8x7B& 8 &7  & Tie\\
5  &Llama-2-7b& 8 & 7  & Mixtral-8x7B& 8 &7  & Tie\\
6  &Llama-2-7b& 8 & 7  & Llama-2-7b& 8 &9  & Llama-2-7b\\
7  &Mixtral-8x7B& 7 & 8  & Llama-2-7b& 7 &8  & Tie\\
8  &Mixtral-8x7B& 7 & 8  & Llama-2-7b& 7 &8  & Tie\\
9  &Mixtral-8x7B& 8 & 9  & Draw & 8 &8  & Mixtral-8x7B\\
10  &Llama-2-7b& 8 & 7  & Mixtral-8x7B& 8 &7  & Tie\\
11  &Llama-2-7b& 8 & 7  & Mixtral-8x7B& 8 &7  & Tie\\
12  &Mixtral-8x7B& 7 & 8  & Llama-2-7b& 8 &9  & Tie\\
13  &Mixtral-8x7B& 7 & 8  & Mixtral-8x7B& 8 &7  & Mixtral-8x7B\\
14  &Mixtral-8x7B& 7 & 8  & Mixtral-8x7B& 8 &7  & Mixtral-8x7B\\
15  &Llama-2-7b& 8 & 7  & Mixtral-8x7B& 8 &7  & Tie\\
16  &Mixtral-8x7B& 8 & 9  & Mixtral-8x7B& 8 &7  & Mixtral-8x7B\\
17  &Mixtral-8x7B& 7 & 8  & Llama-2-7b& 7 &8  & Tie\\
18  &Llama-2-7b& 8 & 7  & Mixtral-8x7B& 8 &7  & Tie\\
19  &Llama-2-7b& 8 & 7  & Mixtral-8x7B& 8 &7  & Tie\\
20  &Mixtral-8x7B& 7 & 8  & Llama-2-7b& 7 &8  & Tie\\
21  &Mixtral-8x7B& 8 & 9  & Mixtral-8x7B& 8 &7  & Mixtral-8x7B\\
22  &Mixtral-8x7B& 6 & 8  & Llama-2-7b& 7 &8  & Tie\\
23  &Mixtral-8x7B& 7 & 8  & Mixtral-8x7B& 8 &7  & Mixtral-8x7B\\
24  &Mixtral-8x7B& 7 & 8  & Mixtral-8x7B& 8 &7  & Mixtral-8x7B\\
25  &Mixtral-8x7B& 8 & 9  & Mixtral-8x7B& 8 &7  & Mixtral-8x7B\\
\end{tabular}
}
\end{table}
\newpage

\subsection{Llama-2-7b vs GPT 3.5}

\begin{table}[h!]
\centering
{\scriptsize
\begin{tabular}{c|ccc|ccc|c}
&\multicolumn{3}{c}{Home}&\multicolumn{3}{c}{Away} & \\[0.2cm]
\midrule       
&  & Side 1& Side 2&  &Side 1 &Side 2  & \\ 
Topic & Winner & Llama-2-7b & GPT-3.5 & Winner & GPT-3.5  & Llama-2-7b  & Overall\\ \midrule 
 1 & Llama-2-7b & 8 & 7    & GPT-3.5  & 9  & 7    & Tie\\
 2 & GPT-3.5 & 7 & 8    & Llama-2-7b  & 7  & 8   & Tie\\
 3 & GPT-3.5 & 8 & 9    & Llama-2-7b  & 8  & 9   & Tie\\
 4 & Llama-2-7b & 8 & 7    & Llama-2-7b & 7  & 8   & Llama-2-7b \\
 5 & GPT-3.5 & 8 & 9   & GPT-3.5 & 8  & 7   & GPT-3.5 \\
 6 & GPT-3.5 & 8 & 9    & Llama-2-7b & 7  & 8   & Tie\\
 7 & GPT-3.5 & 7 & 8    & Llama-2-7b & 7  & 8  & Tie\\
 8 & GPT-3.5 & 6 & 8    & Llama-2-7b & 8  & 7  & Tie\\
 9 & Llama-2-7b & 8 & 7    & GPT-3.5 & 8  & 7   & Tie\\
 10 & GPT-3.5 & 8 & 9   & GPT-3.5 & 8  & 7   & GPT-3.5 \\
 11 & Llama-2-7b & 8 & 7    & GPT-3.5 & 8  & 7   & Tie\\
 12 & GPT-3.5 & 7 & 9    & Llama-2-7b & 7  & 9   & Tie\\
 13 & Llama-2-7b & 8 & 7    & GPT-3.5 & 8  & 7   & Tie\\
 14 & GPT-3.5 & 8 & 9    & GPT-3.5 & 8  & 7   & GPT-3.5 \\
 15 & Llama-2-7b & 8 & 7    & GPT-3.5 & 8  & 7   & Tie\\
 16 & GPT-3.5 & 8 & 9    & Llama-2-7b & 8  & 9   & Tie\\
 17 & GPT-3.5 & 7 & 9    & Llama-2-7b & 7  & 8   & Tie\\
 18 & Llama-2-7b & 8 & 7    & GPT-3.5 & 8  & 7   & Tie\\
 19 & Llama-2-7b & 8 & 7    & GPT-3.5 & 8  & 7   & Tie\\
 20 & GPT-3.5 & 7 & 8   & Llama-2-7b  & 7  & 8   & Tie\\
 21 & GPT-3.5 & 8 & 9   & GPT-3.5 & 8  & 7   & GPT-3.5 \\
 22 & GPT-3.5 & 6 & 8   & Llama-2-7b & 7  & 8   & Tie\\
 23 & GPT-3.5 & 7 & 8   & GPT-3.5 & 8  & 7   & GPT-3.5 \\
 24 & GPT-3.5 & 7 & 8   & Llama-2-7b & 7  & 8   & Tie\\
 25 & GPT-3.5 & 8 & 9   & Llama-2-7b & 8  & 9   & Tie\\
\end{tabular}
}
\end{table}

\subsection{Llama-2-7b vs GPT 4}
\begin{table}[h!]
\centering
{\scriptsize
\begin{tabular}{c|ccc|ccc|c}
&\multicolumn{3}{c}{Home}&\multicolumn{3}{c}{Away} & \\[0.2cm]
\midrule       
&  & Side 1& Side 2&  &Side 1 &Side 2   & \\ 
Topic  & Winner & Llama-2-7b & GPT-4 & Winner & GPT-4  &Llama-2-7b  & Overall\\ \midrule 
 1  &  Llama-2-7b& 9 & 8    & GPT-4& 9 &7    & Tie\\
 2  &  GPT-4& 7 & 9    & GPT-4& 8 &7    & GPT-4\\
 3  &  GPT-4& 8 & 9    & GPT-4& 8 &7    & GPT-4\\
 4  &  GPT-4& 8 & 9    & GPT-4& 8 &7    & GPT-4\\
 5  &  GPT-4& 8 & 9    & GPT-4& 9 &8    & GPT-4\\
 6  &  GPT-4& 7 & 9    & GPT-4& 8 &7    & GPT-4\\
 7  &  GPT-4& 7 & 9    & GPT-4& 8 &7    & GPT-4\\
 8  &  GPT-4& 6 & 8    & GPT-4& 8 &7    & GPT-4\\
 9  &  GPT-4& 7 & 8    & GPT-4& 9 &7    & GPT-4\\
 10 &  GPT-4& 7 & 8    & GPT-4& 8 &7    & GPT-4\\
 11 &  GPT-4& 7 & 8    & GPT-4& 8 &7    & GPT-4\\
 12 &  GPT-4& 6 & 9    & GPT-4& 8 &7    & GPT-4\\
 13 &  GPT-4& 7 & 8    & GPT-4& 9 &7    & GPT-4\\
 14 &  GPT-4& 7 & 9    & GPT-4& 9 &7    & GPT-4\\
 15 &  GPT-4& 7 & 8    & GPT-4& 9 &7    & GPT-4\\
 16 &  GPT-4& 7 & 9    & GPT-4& 8 &7    & GPT-4\\
 17 &  GPT-4& 7 & 9    & GPT-4& 8 &7    & GPT-4\\
 18 &  GPT-4& 8 & 9    & GPT-4& 9 &7    & GPT-4\\
 19 &  GPT-4& 8 & 9    & GPT-4& 8 &7    & GPT-4\\
 20 &  GPT-4& 6 & 8    & GPT-4& 8 &7    & GPT-4\\
 21 &  GPT-4& 7 & 9    & GPT-4& 8 &7    & GPT-4\\
 22 &  GPT-4& 7 & 9    & GPT-4& 8 &7    & GPT-4\\
 23 &  GPT-4& 7 & 8    & GPT-4& 8 &7    & GPT-4\\
 24 &  GPT-4& 7 & 9    & GPT-4& 8 &7    & GPT-4\\
 25 &  GPT-4& 8 & 9   
 & GPT-4& 8 &7    & GPT-4
 \end{tabular}
}
\end{table}
\newpage

\subsection{Llama-2-13b vs Llama-2-13b}
\begin{table}[h!]
\centering
{\scriptsize
\begin{tabular}{c|ccc|ccc}
&\multicolumn{3}{c}{Home}&\multicolumn{3}{c}{Away} \\[0.2cm]
\midrule       
&  & Side 1& Side 2&  &Side 1 &Side 2  \\ 
Topic  & Winner &Llama-2-13b  & Llama-2-13b  & Winner&Llama-2-13b  & Llama-2-13b  \\ \midrule 
1  &Side 1 & 8 & 7  & Side 1 & 8 &7 \\
2  &Side 2 & 8 & 9  & Side 2 & 7 &8 \\
3  &Side 1 & 8 & 7  & Side 1 & 8 &7 \\
4  &Side 1 & 8 & 7  & Side 1 & 8 &7 \\
5  &Side 1 & 8 & 7  & Side 1 & 8 &7 \\
6  &Side 2 & 8 & 9  & Side 1 & 8 &7 \\
7  &Side 1 & 9& 8& Side 1 & 8 &7 \\
8  &Side 2 & 7 & 8  & Side 2 & 7 &8 \\
9  &Side 1 & 8 & 7  & Side 1 & 8 &7 \\
10  &Side 1 & 8 & 7  & Side 1 & 8 &7 \\
11  &Side 2 & 7 & 8  & Side 1 & 8 &7 \\
12  &Side 2 & 7 & 8  & Side 2 & 7 &8 \\
13  &Side 1 & 8 & 7  & Side 1 & 8 &7 \\
14  &Side 1 & 8 & 7  & Side 1 & 8 &7 \\
15  &Side 1 & 8 & 7  & Side 1 & 8 &7 \\
16  &Side 1 & 8 & 8  & Side 1 & 8 &8 \\
17  &Side 2 & 7 & 8  & Side 2 & 7 &8 \\
18  &Side 1 & 8 & 7  & Side 1 & 8 &7 \\
19  &Side 1 & 9 & 7  & Side 1 & 9 &7 \\
20  &Side 2 & 6 & 8  & Side 2 & 7 &8 \\
21  &Side 2 & 8 & 9  & Side 2 & 8 &9 \\
22  &Side 2 & 6 & 8  & Side 2 & 6 &8 \\
23  &Side 1 & 8 & 7  & Side 1 & 8 &7 \\
24  &Side 2 & 8 & 9  & Side 2 & 8 &9\\
25  &Side 2 & 8 & 9  & Side 2 & 8 &9 \\
\end{tabular}
}
\end{table}

\subsection{Llama-2-13b vs Llama-2-70b}

\begin{table}[h!]
\centering
{\scriptsize
\begin{tabular}{c|ccc|ccc|c}
&\multicolumn{3}{c}{Home}&\multicolumn{3}{c}{Away} & \\[0.2cm]
\midrule       
&  & Side 1& Side 2&  &Side 1 &Side 2   & \\ 
Topic  & Winner &Llama-2-13b  & Llama-2-70b  & Winner&Llama-2-70b  & Llama-2-13b   & Overall\\ \midrule 
1  &Llama-2-13b& 8 & 6  & Llama-2-70b& 8 &7  & Tie\\
2  &Llama-2-70b& 7 & 8  & Llama-2-13b& 8 &9  & Tie\\
3  &Llama-2-70b& 8 & 9  & Llama-2-70b& 8 &7  & Llama-2-70b\\
4  &Llama-2-13b& 8 & 7  & Llama-2-70b& 8 &7  & Tie\\
5  &Llama-2-13b& 8 & 7  & Llama-2-70b& 8 &7  & Tie\\
6  &Llama-2-70b& 8 & 9  & Llama-2-70b& 8 &7  & Llama-2-70b\\
7  &Llama-2-70b& 7 & 8  & Llama-2-70b& 8 &7  & Llama-2-70b\\
8  &Llama-2-70b& 7 & 8  & Llama-2-70b& 8 &7  & Llama-2-70b\\
9  &Llama-2-70b& 7 & 8  & Llama-2-70b& 8 &7  & Llama-2-70b\\
10  &Llama-2-13b& 8 & 7  & Llama-2-70b& 8 &7  & Tie\\
11  &Llama-2-13b& 8 & 7  & Llama-2-70b& 8 &7  & Tie\\
12  &Llama-2-70b& 7 & 8  & Llama-2-13b& 8 &9  & Tie\\
13  &Llama-2-13b& 8 & 7  & Llama-2-70b& 8 &7  & Tie\\
14  &Llama-2-13b& 8 & 7  & Llama-2-70b& 8 &7  & Tie\\
15  &Llama-2-13b& 8 & 6  & Llama-2-70b& 8 &6  & Tie\\
16  &Llama-2-70b& 8 & 9  & Llama-2-70b& 8 &7  & Llama-2-70b\\
17  &Llama-2-70b& 7 & 8  & Llama-2-13b& 7 &8  & Tie\\
18  &Llama-2-13b& 8 & 7  & Llama-2-70b& 8 &7  & Tie\\
19  &Llama-2-13b& 8 & 6  & Llama-2-70b& 8 &6  & Tie\\
20  &Llama-2-70b& 6 & 8  & Llama-2-13b& 7 &8  & Tie\\
21  &Llama-2-70b& 8 & 9  & Llama-2-70b& 8 &7  & Llama-2-70b\\
22  &Llama-2-70b& 6 & 8  & Llama-2-13b& 7 &8  & Tie\\
23  &Llama-2-13b& 8 & 7  & Llama-2-70b& 8 &7  & Tie\\
24  &Llama-2-70b& 7 & 8  & Llama-2-13b& 8 &9  & Tie\\
25  &Llama-2-13b& 8 & 7  & Llama-2-70b& 8 &8  & Tie\\
\end{tabular}
}
\end{table}

\newpage 

\subsection{Llama-2-13b vs. Llama-3-70b}
\begin{table}[h!]
\centering
{\scriptsize
\begin{tabular}{c|ccc|ccc|c}
&\multicolumn{3}{c}{Home}&\multicolumn{3}{c}{Away} & \\[0.2cm]
\midrule       
&  & Side 1& Side 2&  &Side 1 &Side 2   &\\ 
Topic  & Winner &Llama-3-70b& Llama-2-13b& Winner&Llama-2-13b& Llama-3-70b&Overall\\ \midrule 
1  &Llama-3-70b& 9& 7& Llama-2-13b& 8&7&Tie\\
2  &tie& 8& 8& Llama-3-70b& 7&8&Llama-3-70b\\
3  &Llama-3-70b& 8& 7& Llama-3-70b& 8&9&Llama-3-70b\\
4  &Llama-3-70b& 8& 7& Llama-2-13b& 8&7&Tie\\
5  &Llama-3-70b& 8& 7& Llama-2-13b& 8&7&Tie\\
6  &Llama-2-13b& 8& 9& Llama-2-13b& 8&7&Llama-2-13b\\
7  &Llama-3-70b& 8& 7& Llama-3-70b& 8&9&Llama-3-70b\\
8  &Llama-2-13b& 7& 8& Llama-3-70b& 6&8&Tie\\
9  &Llama-3-70b& 8& 7& Llama-3-70b& 8&9&Llama-3-70b\\
10  &Llama-3-70b& 8& 7& Llama-2-13b& 8&7&Llama-2-13b\\
11  &Llama-3-70b& 8& 6& Llama-3-70b& 7&8&Llama-3-70b\\
12  &Llama-3-70b& 8& 7& Llama-3-70b& 7&9&Llama-3-70b\\
13  &Llama-3-70b& 8& 6& Llama-3-70b& 7&8&Llama-3-70b\\
14  &Llama-3-70b& 8& 7& Llama-2-13b& 8&7&Tie\\
15  &Llama-3-70b& 9& 7& Llama-2-13b& 8&7&Tie\\
16  &Llama-3-70b& 8& 7& Llama-3-70b& 8&9&Llama-3-70b\\
17  &Llama-2-13b& 7& 8& Llama-3-70b& 7&9&Tie\\
18  &Llama-3-70b& 9& 8& Llama-3-70b& 7&8&Llama-3-70b\\
19  &Llama-3-70b& 9& 7& Llama-2-13b& 8&7&Tie\\
20  &Llama-2-13b& 7& 8& Llama-3-70b& 6&8&Tie\\
21  &Llama-3-70b& 9& 8& Llama-3-70b& 7&8&Llama-3-70b\\
22  &Llama-2-13b& 6& 8& Llama-3-70b& 6&8&Ties\\
23  &Llama-3-70b& 8& 7& Llama-3-70b& 7&8&Llama-3-70b\\
24  &Llama-3-70b& 8& 7& Llama-3-70b& 7&8&Llama-3-70b\\
25  &Llama-3-70b& 8& 7& Llama-3-70b& 8&9&Llama-3-70b\\
\end{tabular}
}
\end{table}

\subsection{Llama-2-13b vs. Vicuna-7b-v1.5}

\begin{table}[h!]
\centering
{\scriptsize
\begin{tabular}{c|ccc|ccc|c}
&\multicolumn{3}{c}{Home}&\multicolumn{3}{c}{Away} & \\[0.2cm]
\midrule       
&  & Side 1& Side 2&  &Side 1 &Side 2   & \\ 
Topic  & Winner &Llama-2-13b  & Vicuna-7b-v1.5  & Winner&Vicuna-7b-v1.5  & Llama-2-13b   & Overall\\ \midrule 
1  &Llama-2-13b& 8 & 7  & Vicuna-7b-v1.5& 8 &7  & Tie\\
2  &Vicuna-7b-v1.5& 7 & 8  & Llama-2-13b& 7 &8  & Tie\\
3  &Llama-2-13b& 9 & 8  & Vicuna-7b-v1.5& 8 &8  & Tie\\
4  &Llama-2-13b& 8 & 7  & Vicuna-7b-v1.5& 8 &7  & Tie\\
5  &Llama-2-13b& 8 & 7  & Vicuna-7b-v1.5& 8 &7  & Tie\\
6  &Llama-2-13b& 8 & 7  & Llama-2-13b& 7 &8  & Llama-2-13b\\
7  &Llama-2-13b& 8 & 7  & Llama-2-13b& 7 &8  & Llama-2-13b\\
8  &Llama-2-13b& 8 & 7  & Llama-2-13b& 7 &8  & Llama-2-13b\\
9  &Llama-2-13b& 8 & 7  & Vicuna-7b-v1.5& 8 &7  & Tie\\
10  &Llama-2-13b& 8 & 7  & Llama-2-13b& 7 &8  & Llama-2-13b\\
11  &Llama-2-13b& 8 & 7  & Llama-2-13b& 7 &8  & Llama-2-13b\\
12  &Llama-2-13b& 8 & 7  & Llama-2-13b& 6 &8  & Llama-2-13b\\
13  &Llama-2-13b& 9 & 7  & Llama-2-13b& 7 &8  & Llama-2-13b\\
14  &Llama-2-13b& 8 & 7  & Vicuna-7b-v1.5& 8 &7  & Tie\\
15  &Llama-2-13b& 8 & 6  & Vicuna-7b-v1.5& 8 &7  & Tie\\
16  &Llama-2-13b& 8 & 7  & Llama-2-13b& 7 &8  & Llama-2-13b\\
17  &Llama-2-13b& 8 & 7  & Llama-2-13b& 6 &9  & Llama-2-13b\\
18  &Llama-2-13b& 8 & 6  & Vicuna-7b-v1.5& 8 &7  & Tie\\
19  &Llama-2-13b& 8 & 7  & Llama-2-13b& 7 &8  & Llama-2-13b\\
20  &Vicuna-7b-v1.5& 6 & 8  & Llama-2-13b& 6 &8  & Tie\\
21  &Llama-2-13b& 8 & 7  & Llama-2-13b& 6 &8  & Llama-2-13b\\
22  &Vicuna-7b-v1.5& 6 & 8  & Llama-2-13b& 6 &8  & Tie\\
23  &Llama-2-13b& 8 & 7  & Llama-2-13b& 7 &8  & Llama-2-13b\\
24  &Llama-2-13b& 8 & 7  & Llama-2-13b& 8 &9  & Llama-2-13b\\
25  &Llama-2-13b& 8 & 7  & Llama-2-13b& 7 &8  & Llama-2-13b\\
\end{tabular}
}
\end{table}

\newpage

\subsection{Llama-2-13b vs. Vicuna-13b-v1.5}

\begin{table}[h!]
\centering
{\scriptsize
\begin{tabular}{c|ccc|ccc|c}
&\multicolumn{3}{c}{Home}&\multicolumn{3}{c}{Away} & \\[0.2cm]
\midrule       
&  & Side 1& Side 2&  &Side 1 &Side 2   & \\ 
Topic  & Winner &Llama-2-13b  & Vicuna-13b-v1.5  & Winner&Vicuna-13b-v1.5  & Llama-2-13b   & Overall\\ \midrule 
1  &Llama-2-13b& 8 & 7  & Vicuna-13b-v1.5& 8 &7  & Tie\\
2  &Vicuna-13b-v1.5& 8 & 9  & Llama-2-13b& 7 &8  & Tie\\
3  &Llama-2-13b& 8 & 7  & Llama-2-13b& 7 &8  & Llama-2-13b\\
4  &Llama-2-13b& 8 & 7  & Llama-2-13b& 7 &8  & Llama-2-13b\\
5  &Llama-2-13b& 8 & 7  & Vicuna-13b-v1.5& 8 &7  & Tie\\
6  &Vicuna-13b-v1.5& 7 & 8  & Llama-2-13b& 7 &8  & Tie\\
7  &Vicuna-13b-v1.5& 8 & 9  & Llama-2-13b& 7 &8  & Tie\\
8  &Vicuna-13b-v1.5& 7 & 8  & Llama-2-13b& 6 &8  & Tie\\
9  &Llama-2-13b& 8 & 7  & Vicuna-13b-v1.5& 8 &7  & Tie\\
10  &Llama-2-13b& 8 & 7  & Llama-2-13b& 7 &8  & Llama-2-13b\\
11  &Llama-2-13b& 8 & 7  & Vicuna-13b-v1.5& 8 &7  & Tie\\
12  &Vicuna-13b-v1.5& 8 & 9  & Llama-2-13b& 6 &8  & Tie\\
13  &Llama-2-13b& 8 & 6  & Vicuna-13b-v1.5& 8 &7  & Tie\\
14  &Llama-2-13b& 8 & 7  & Vicuna-13b-v1.5& 8 &7  & Tie\\
15  &Llama-2-13b& 8 & 6  & Vicuna-13b-v1.5& 8 &7  & Tie\\
16  &Llama-2-13b& 8 & 7  & Llama-2-13b& 7 &8  & Llama-2-13b\\
17  &Vicuna-13b-v1.5& 7 & 8  & Llama-2-13b& 6 &8  & Tie\\
18  &Llama-2-13b& 8 & 7  & Vicuna-13b-v1.5& 8 &7  & Tie\\
19  &Llama-2-13b& 9 & 7  & Llama-2-13b& 7 &8  & Llama-2-13b\\
20  &Vicuna-13b-v1.5& 7 & 8  & Llama-2-13b& 6 &8  & Tie\\
21  &Llama-2-13b& 8 & 7  & Llama-2-13b& 7 &8  & Llama-2-13b\\
22  &Vicuna-13b-v1.5& 6 & 8  & Llama-2-13b& 7 &8  & Tie\\
23  &Llama-2-13b& 8 & 7  & Llama-2-13b& 7 &8  & Llama-2-13b\\
24  &Llama-2-13b& 8 & 7  & Vicuna-13b-v1.5& 8 &7  & Tie\\
25  &Llama-2-13b& 8 & 7  & Llama-2-13b& 8 &9  & Llama-2-13b\\
\end{tabular}
}
\end{table}

\subsection{Llama-2-13b vs Mixtral-8x7B}

\begin{table}[h!]
\centering
{\scriptsize
\begin{tabular}{c|ccc|ccc|c}
&\multicolumn{3}{c}{Home}&\multicolumn{3}{c}{Away} & \\[0.2cm]
\midrule       
&  & Side 1& Side 2&  &Side 1 &Side 2   &\\ 
Topic  & Winner &Llama-2-13b  & Mixtral-8x7B  & Winner&Mixtral-8x7B  & Llama-2-13b   &Overall\\ \midrule 
1  &Llama-2-13b& 8 & 7  & Mixtral-8x7B& 9 &7  &Tie\\
2  &Mixtral-8x7B& 7 & 8  & Llama-2-13b& 8 &9  &Tie\\
3  &Llama-2-13b& 8 & 7  & Mixtral-8x7B& 8 &7  &Tie\\
4  &Mixtral-8x7B& 8 & 9  & Mixtral-8x7B& 8 &7  &Mixtral-8x7B\\
5  &Llama-2-13b& 8 & 7  & Mixtral-8x7B& 8 &7  &Tie\\
6  &Mixtral-8x7B& 8 & 9  & Llama-2-13b& 7 &8  &Tie\\
7  &Mixtral-8x7B& 8 & 9  & Llama-2-13b& 7 &8  &Tie\\
8  &Mixtral-8x7B& 7 & 8  & Llama-2-13b& 6 &8  &Tie\\
9  &Llama-2-13b& 8 & 7  & Mixtral-8x7B& 8 &7  &Tie\\
10  &Llama-2-13b& 8 & 7  & Mixtral-8x7B& 8 &7  &Tie\\
11  &Llama-2-13b& 8 & 7  & Mixtral-8x7B& 8 &7  &Tie\\
12  &Mixtral-8x7B& 6 & 8  & Llama-2-13b& 8 &9  &Tie\\
13  &Llama-2-13b& 8 & 7  & Mixtral-8x7B& 8 &7  &Tie\\
14  &Llama-2-13b& 8 & 7  & Mixtral-8x7B& 8 &7  &Tie\\
15  &Llama-2-13b& 8 & 6  & Mixtral-8x7B& 8 &7  &Tie\\
16  &Llama-2-13b& 8 & 7  & Mixtral-8x7B& 8 &7  &Tie\\
17  &Mixtral-8x7B& 7 & 8  & Llama-2-13b& 7 &8  &Tie\\
18  &Llama-2-13b& 8 & 7  & Mixtral-8x7B& 8 &7  &Tie\\
19  &Llama-2-13b& 8 & 7  & Mixtral-8x7B& 8 &7  &Tie\\
20  &Mixtral-8x7B& 6 & 8  & Llama-2-13b& 6 &9  &Tie\\
21  &Mixtral-8x7B& 8 & 9  & Mixtral-8x7B& 8 &7  &Mixtral-8x7B\\
22  &Mixtral-8x7B& 6 & 8  & Llama-2-13b& 6 &8  &Tie\\
23  &Mixtral-8x7B& 7 & 8  & Mixtral-8x7B& 8 &7  &Mixtral-8x7B\\
24  &Mixtral-8x7B& 7 & 8  & Mixtral-8x7B& 8 &7  &Mixtral-8x7B\\
25  &Mixtral-8x7B& 7 & 8  & Mixtral-8x7B& 8 &7  &Mixtral-8x7B\\
\end{tabular}
}
\end{table}

\newpage
\subsection{Llama-2-13b vs. GPT 3.5}

\begin{table}[h!]
\centering
{\scriptsize
\begin{tabular}{c|ccc|ccc|c}
&\multicolumn{3}{c}{Home}&\multicolumn{3}{c}{Away} & \\[0.2cm]
\midrule       
&  & Side 1& Side 2&  &Side 1 &Side 2   & \\ 
Topic  & Winner &Llama-2-13b& GPT 3.5& Winner&GPT 3.5& Llama-2-13b & Overall\\ \midrule 
1  &Llama-2-13b & 8& 7& GPT 3.5& 9&7 & Tie\\
2  &GPT 3.5& 7& 9& Llama-2-13b & 8&9 & Tie\\
3  &Llama-2-13b & 8& 7& GPT 3.5& 8&7 & Tie\\
4  &Llama-2-13b & 8& 7& GPT 3.5& 8&7 & Tie\\
5  &GPT 3.5& 8& 9& GPT 3.5& 8&7 & GPT 3.5\\
6  &GPT 3.5& 8& 9& Llama-2-13b & 7&8 & Tie\\
7  &GPT 3.5& 8& 9& Llama-2-13b & 7&8 & Tie\\
8  &GPT 3.5& 7& 9& GPT 3.5& 8&7 & GPT 3.5\\
9  &Llama-2-13b & 8& 7& GPT 3.5& 8&7 & Tie\\
10  &Llama-2-13b & 8& 7& GPT 3.5& 8&7 & Tie\\
11  &GPT 3.5& 8& 9& Llama-2-13b & 7&8 & Tie\\
12  &GPT 3.5& 8& 9& Llama-2-13b & 7&8 & Tie\\
13  &GPT 3.5& 7& 8& GPT 3.5& 8&7 & GPT 3.5\\
14  &tie& 8& 8& GPT 3.5& 8&7 & GPT 3.5\\
15  &Llama-2-13b & 8& 7& GPT 3.5& 9&7 & Tie\\
16  &GPT 3.5& 8& 9& GPT 3.5& 8&7 & GPT 3.5\\
17  &GPT 3.5& 6& 8& Llama-2-13b & 8&9 & Tie\\
18  &Llama-2-13b & 8& 7& GPT 3.5& 8&7 & Tie\\
19  &Llama-2-13b & 8& 7& GPT 3.5& 8&7 & Tie\\
20  &GPT 3.5& 6& 8& Llama-2-13b & 6&9 & Tie\\
21  &GPT 3.5& 8& 9& Llama-2-13b & 7&8 & Tie\\
22  &GPT 3.5& 6& 8& Llama-2-13b & 7&8 & Tie\\
23  &GPT 3.5& 8& 9& GPT 3.5& 8&7 & GPT 3.5\\
24  &GPT 3.5& 7& 8& Llama-2-13b & 8&9 & Tie\\
25  &GPT 3.5& 7& 8& GPT 3.5& 8&7 & GPT 3.5\\
\end{tabular}
}
\end{table}

\subsection{Llama-2-13b vs. GPT 4}
\begin{table}[h!]
\centering
{\scriptsize
\begin{tabular}{c|ccc|ccc|c}
&\multicolumn{3}{c}{Home}&\multicolumn{3}{c}{Away} & \\[0.2cm]
\midrule       
&  & Side 1& Side 2&  &Side 1 &Side 2  &\\ 
Topic& Winner & Llama-2-13b & GPT-4 & Winner &GPT-4  &Llama-2-13b   &Overall\\ \midrule 
1 & GPT-4 & 8 & 9    & GPT-4  &9  &6      &GPT-4  \\
 2 & GPT-4 & 6 & 9    & Llama-2-13b  &8  &9    &Tie\\
 3 & GPT-4 & 8 & 9    & GPT-4  &9  &8    &GPT-4  \\
 4 & GPT-4 & 7 & 8    & GPT-4 &8  &7    &GPT-4  \\
 5 & GPT-4 & 8 & 9    & GPT-4 &9  &8    &GPT-4  \\
 6 & GPT-4 & 7 & 9    & GPT-4 &8  &7    &GPT-4  \\
 7 & GPT-4 & 7 & 9    & GPT-4 &8  &7    &GPT-4  \\
 8 & GPT-4 & 7 & 8    & GPT-4 &8  &7   &GPT-4  \\
 9 & GPT-4 & 8 & 9    & GPT-4 &8  &7    &GPT-4  \\
 10 & GPT-4 & 8 & 9    & GPT-4 &8  &7    &GPT-4  \\
 11 & GPT-4 & 7 & 8    & GPT-4 &9  &7    &GPT-4  \\
 12 & GPT-4 & 7 & 9    & Llama-2-13b &8  &9    &Tie\\
 13 & GPT-4 & 7 & 8    & GPT-4 &8  &7    &GPT-4  \\
 14 & GPT-4 & 7 & 9    & GPT-4 &9  &8    &GPT-4  \\
 15 & GPT-4 & 8 & 9    & GPT-4 &9  &7    &GPT-4  \\
 16 & GPT-4 & 8 & 9    & GPT-4 &8  &7    &GPT-4  \\
 17 & GPT-4 & 6 & 9    & GPT-4 &8  &7    &GPT-4  \\
 18 & GPT-4 & 8 & 9    & GPT-4 &9  &7    &GPT-4  \\
 19 & GPT-4 & 8 & 9    & GPT-4 &8  &7    &GPT-4  \\
 20 & GPT-4 & 6 & 9   & Llama-2-13b  &8  &9    &Tie\\
 21 & GPT-4 & 7 & 9   & GPT-4 &8  &7    &GPT-4  \\
 22 & GPT-4 & 5 & 8   & GPT-4 &8  &7    &GPT-4  \\
 23 & GPT-4 & 7 & 8   & GPT-4 &8  &7    &GPT-4  \\
 24 & GPT-4 & 7 & 8   & GPT-4 &8  &7    &GPT-4  \\
 25 & GPT-4 & 8 & 9   & GPT-4 &8  &7    &GPT-4  
\end{tabular}
}
\end{table}
\newpage

\subsection{Llama-2-70b vs. Llama-2-70b}

\begin{table}[h!]
\centering
{\scriptsize
\begin{tabular}{c|ccc|ccc}
&\multicolumn{3}{c}{Home}&\multicolumn{3}{c}{Away} \\[0.2cm]
\midrule       
&  & Side 1& Side 2&  &Side 1 &Side 2  \\ 
Topic  & Winner &Llama-2-70b& Llama-2-70b& Winner&Llama-2-70b& Llama-2-70b\\ \midrule 
1  &1& 9& 7& 1& 8&7\\
2  &2& 7& 8& 2& 7&8\\
3  &2& 8& 9& 2& 8&9\\
4  &1& 8& 7& 1& 8&7\\
5  &1& 8& 7& 1& 8&7\\
6  &2& 8& 9& 2& 8&9\\
7  &2& 8& 9& 2& 8&9\\
8  &2& 8& 9& 1& 8&7\\
9  &1& 9& 7& 1& 8&7\\
10  &2& 8& 9& 2& 8&9\\
11  &1& 8& 7& 1& 8&7\\
12  &2& 8& 9& 2& 8&9\\
13  &1& 8& 7& 1& 9&7\\
14  &1& 8& 7& 1& 8&7\\
15  &1& 8& 7& 1& 9&7\\
16  &1& 8& 7& 1& 9&8\\
17  &2& 7& 8& 2& 8&9\\
18  &1& 8& 7& 1& 8&7\\
19  &1& 8& 7& 1& 8&7\\
20  &2& 6& 8& 2& 7&8\\
21  &1& 8& 7& 2& 8&9\\
22  &2& 7& 8& 2& 7&8\\
23  &1& 8& 7& 1& 8&7\\
24  &1& 8& 7& 1& 8&7\\
25  &2& 8& 9& 2& 8&9\\
\end{tabular}
}
\end{table}

\subsection{Llama-2-70b vs. Llama-3-70b}

\begin{table}[h!]
\centering
{\scriptsize
\begin{tabular}{c|ccc|ccc|c}
&\multicolumn{3}{c}{Home}&\multicolumn{3}{c}{Away} & \\[0.2cm]
\midrule       
&  & Side 1& Side 2&  &Side 1 &Side 2   &\\ 
Topic  & Winner &Llama-3-70b& Llama-2-70b& Winner&Llama-2-70b& Llama-3-70b&Overall\\ \midrule 
1  &Llama-3-70b& 9& 7& Llama-2-70b& 8&7&Tie\\
2  &Llama-2-70b& 8& 9& Llama-3-70b& 7&9&Tie\\
3  &Llama-3-70b& 8& 7& Llama-3-70b& 7&8&Llama-3-70b\\
4  &Llama-3-70b& 8& 7& Llama-3-70b& 7&8&Llama-3-70b\\
5  &Llama-3-70b& 8& 7& Llama-3-70b& 8&9&Llama-3-70b\\
6  &Llama-2-70b& 8& 9& Llama-3-70b& 7&8&Tie\\
7  &Llama-2-70b& 7& 9& Llama-3-70b& 7&8&Tie\\
8  &Llama-2-70b& 7& 8& Llama-3-70b& 7&8&Tie\\
9  &Llama-3-70b& 8& 7& Llama-2-70b& 8&7&Tie\\
10  &Llama-3-70b& 8& 7& Llama-3-70b& 7&8&Llama-3-70b\\
11  &Llama-3-70b& 8& 7& Llama-3-70b& 7&8&Llama-3-70b\\
12  &Llama-2-70b& 8& 9& Llama-3-70b& 7&8&Tie\\
13  &Llama-3-70b& 9& 7& Llama-3-70b& 8&9&Llama-3-70b\\
14  &Llama-3-70b& 8& 7& Llama-2-70b& 8&7&Tie\\
15  &Llama-3-70b& 9& 6& Llama-2-70b& 9&8&Tie\\
16  &Llama-2-70b& 8& 9& Llama-2-70b& 8&7&Tie\\
17  &Llama-2-70b& 7& 8& Llama-3-70b& 7&8&Tie\\
18  &Llama-3-70b& 8& 7& Llama-2-70b& 8&7&Tie\\
19  &Llama-3-70b& 8& 7& Llama-2-70b& 8&7&Tie\\
20  &Llama-2-70b& 8& 9& Llama-3-70b& 7&9&Tie\\
21  &Tie& 8& 8& Llama-3-70b& 8&9&Llama-3-70b\\
22  &Llama-2-70b& 7& 8& Llama-3-70b& 6&8&Tie\\
23  &Llama-3-70b& 8& 7& Llama-2-70b& 8&7&Tie\\
24  &Llama-3-70b& 8& 7& Llama-3-70b& 7&8&Llama-3-70b\\
25  &Llama-3-70b& 8& 7& Llama-3-70b& 8&9&Llama-3-70b\\
\end{tabular}
}
\end{table}

\newpage

\subsection{Llama-2-70b vs. Vicuna-7b-v1.5}

\begin{table}[h!]
\centering
{\scriptsize
\begin{tabular}{c|ccc|ccc|c}
&\multicolumn{3}{c}{Home}&\multicolumn{3}{c}{Away} & \\[0.2cm]
\midrule       
&  & Side 1& Side 2&  &Side 1 &Side 2   &\\ 
Topic  & Winner &Llama-2-70b& Vicuna-7b-v1.5& Winner&Vicuna-7b-v1.5& Llama-2-70b &Overall\\ \midrule 
1  &Llama-2-70b& 9& 7& Vicuna-7b-v1.5& 8&7 &Tie\\
2  &Vicuna-7b-v1.5& 8& 9& Llama-2-70b& 7&8 &Tie\\
3  &Llama-2-70b& 9& 8& Llama-2-70b& 8&9 &Llama-2-70b\\
4  &Llama-2-70b& 8& 7& Llama-2-70b& 7&8 &Llama-2-70b\\
5  &Llama-2-70b& 9& 8& Vicuna-7b-v1.5& 8&7 &Tie\\
6  &Llama-2-70b& 8& 7& Llama-2-70b& 8&9 &Llama-2-70b\\
7  &Llama-2-70b& 8& 7& Llama-2-70b& 7&8 &Llama-2-70b\\
8  &Llama-2-70b& 8& 7& Llama-2-70b& 7&8 &Llama-2-70b\\
9  &Llama-2-70b& 8& 7& Llama-2-70b& 7&9 &Llama-2-70b\\
10  &Llama-2-70b& 8& 7& Llama-2-70b& 7&8 &Llama-2-70b\\
11  &Llama-2-70b& 8& 7& Vicuna-7b-v1.5& 8&7 &Tie\\
12  &Llama-2-70b& 8& 7& Llama-2-70b& 6&8 &Llama-2-70b\\
13  &Llama-2-70b& 8& 7& Llama-2-70b& 7&8 &Llama-2-70b\\
14  &Llama-2-70b& 9& 7& Vicuna-7b-v1.5& 8&7 &Tie\\
15  &Llama-2-70b& 8& 6& Vicuna-7b-v1.5& 8&7 &Tie\\
16  &Llama-2-70b& 9& 8& Llama-2-70b& 7&9 &Llama-2-70b\\
17  &Vicuna-7b-v1.5& 7& 8& Llama-2-70b& 7&8 &Tie\\
18  &Llama-2-70b& 9& 7& Vicuna-7b-v1.5& 8&7 &Tie\\
19  &Llama-2-70b& 9& 7& Vicuna-7b-v1.5& 8&7 &Tie\\
20  &Vicuna-7b-v1.5& 7& 8& Llama-2-70b& 6&8 &Tie\\
21  &Llama-2-70b& 8& 7& Llama-2-70b& 8&9 &Llama-2-70b\\
22  &Vicuna-7b-v1.5& 7& 8& Llama-2-70b& 5&8 &Tie\\
23  &Llama-2-70b& 8& 7& Llama-2-70b& 6&8 &Llama-2-70b\\
24  &Llama-2-70b& 8& 7& Vicuna-7b-v1.5& 8&7 &Tie\\
25  &Llama-2-70b& 8& 7& Llama-2-70b& 7&8 &Llama-2-70b\\
\end{tabular}
}
\end{table}

\subsection{Llama-2-70b vs. Vicuna-13b-v1.5}

\begin{table}[h!]
\centering
{\scriptsize
\begin{tabular}{c|ccc|ccc|c}
&\multicolumn{3}{c}{Home}&\multicolumn{3}{c}{Away} & \\[0.2cm]
\midrule       
&  & Side 1& Side 2&  &Side 1 &Side 2   &\\ 
Topic  & Winner &Llama-2-70b& Vicuna-13b-v1.5& Winner&Vicuna-13b-v1.5& Llama-2-70b &Overall\\ \midrule 
1  &Llama-2-70b& 8& 7& Vicuna-13b-v1.5& 8&7 &Tie\\
2  &Vicuna-13b-v1.5& 7& 8& Llama-2-70b& 7&8 &Tie\\
3  &Llama-2-70b& 8& 7& Llama-2-70b& 7&8 &Llama-2-70b\\
4  &Llama-2-70b& 8& 7& Llama-2-70b& 7&8 &Llama-2-70b\\
5  &Llama-2-70b& 8& 7& Llama-2-70b& 7&8 &Llama-2-70b\\
6  &Llama-2-70b& 8& 7& Llama-2-70b& 7&9 &Llama-2-70b\\
7  &Llama-2-70b& 8& 7& Llama-2-70b& 7&8 &Llama-2-70b\\
8  &Vicuna-13b-v1.5& 7& 8& Llama-2-70b& 6&8 &Tie\\
9  &Llama-2-70b& 8& 7& Vicuna-13b-v1.5& 8&7 &Tie\\
10  &Llama-2-70b& 8& 7& Llama-2-70b& 7&8 &Llama-2-70b\\
11  &Llama-2-70b& 8& 7& Vicuna-13b-v1.5& 8&7 &Tie\\
12  &Vicuna-13b-v1.5& 8& 9& Llama-2-70b& 6&9 &Tie\\
13  &Llama-2-70b& 8& 7& Llama-2-70b& 7&8 &Llama-2-70b\\
14  &Llama-2-70b& 8& 7& Vicuna-13b-v1.5& 8&7 &Tie\\
15  &Llama-2-70b& 9& 7& Vicuna-13b-v1.5& 8&7 &Tie\\
16  &Llama-2-70b& 8& 7& Llama-2-70b& 7&8 &Llama-2-70b\\
17  &Vicuna-13b-v1.5& 7& 8& Llama-2-70b& 7&8 &Tie\\
18  &Llama-2-70b& 8& 7& Vicuna-13b-v1.5& 8&7 &Tie\\
19  &Llama-2-70b& 9& 7& Vicuna-13b-v1.5& 8&7 &Tie\\
20  &Vicuna-13b-v1.5& 7& 8& Llama-2-70b& 6&8 &Tie\\
21  &Vicuna-13b-v1.5& 8& 9& Llama-2-70b& 7&8 &Tie\\
22  &Llama-2-70b& 8& 7& Llama-2-70b& 6&7 &Llama-2-70b\\
23  &Llama-2-70b& 8& 7& Llama-2-70b& 7&8 &Llama-2-70b\\
24  &Llama-2-70b& 8& 7& Llama-2-70b& 8&9 &Llama-2-70b\\
25  &Vicuna-13b-v1.5& 7& 8& Llama-2-70b& 7&8 &Tie\\
\end{tabular}
}
\end{table}
\newpage

\subsection{Llama-2-70b vs. Mixtral-8x7B}

\begin{table}[h!]
\centering
{\scriptsize
\begin{tabular}{c|ccc|ccc|c}
&\multicolumn{3}{c}{Home}&\multicolumn{3}{c}{Away} & \\[0.2cm]
\midrule       
&  & Side 1& Side 2&  &Side 1 &Side 2   &\\ 
Topic  & Winner &Llama-2-70b  & Mixtral-8x7B  & Winner&Mixtral-8x7B  & Llama-2-70b   &Overall\\ \midrule 
1  &Llama-2-70b& 9 & 7  & Mixtral-8x7B& 8 &7  &Tie\\
2  &Mixtral-8x7B& 8 & 9  & Llama-2-70b& 7 &8  &Tie\\
3  &Llama-2-70b& 8 & 7  & Mixtral-8x7B& 8 &7  &Tie\\
4  &Llama-2-70b& 8 & 7  & Mixtral-8x7B& 8 &7  &Tie\\
5  &Llama-2-70b& 8 & 7  & Mixtral-8x7B& 8 &7  &Tie\\
6  &Mixtral-8x7B& 8 & 9  & Llama-2-70b& 8 &9  &Tie\\
7  &Llama-2-70b& 8 & 7  & Llama-2-70b& 8 &9  &Llama-2-70b\\
8  &Mixtral-8x7B& 8 & 9  & Llama-2-70b& 7 &8  &Tie\\
9  &Llama-2-70b& 8 & 7  & Mixtral-8x7B& 8 &7  &Tie\\
10  &Llama-2-70b& 8 & 7  & Mixtral-8x7B& 8 &7  &Tie\\
11  &Llama-2-70b& 8 & 7  & Mixtral-8x7B& 8 &7  &Tie\\
12  &Mixtral-8x7B& 8 & 9  & Llama-2-70b& 8 &9  &Tie\\
13  &Mixtral-8x7B& 7 & 8  & Mixtral-8x7B& 8 &7  &Mixtral-8x7B\\
14  &tie & 8 & 8  & Mixtral-8x7B& 8 &7  &Mixtral-8x7B\\
15  &Llama-2-70b& 8 & 7  & Mixtral-8x7B& 8 &7  &Mixtral-8x7B\\
16  &Mixtral-8x7B& 8 & 9  & Llama-2-70b& 8 &9  &Tie\\
17  &Mixtral-8x7B& 7 & 8  & Llama-2-70b& 7 &8  &Tie\\
18  &Llama-2-70b& 8 & 7  & Mixtral-8x7B& 8 &7  &Tie\\
19  &Llama-2-70b& 8 & 7  & Mixtral-8x7B& 8 &7  &Tie\\
20  &Mixtral-8x7B& 7 & 8  & Llama-2-70b& 7 &9  &Tie\\
21  &tie & 8 & 8  & Llama-2-70b& 8 &9  &Llama-2-70b\\
22  &Mixtral-8x7B& 6 & 8  & Llama-2-70b& 6 &8  &Tie\\
23  &Llama-2-70b& 8 & 7  & Mixtral-8x7B& 8 &7  &Tie\\
24  &Llama-2-70b& 8 & 7  & Mixtral-8x7B& 8 &7  &Tie\\
25  &Mixtral-8x7B& 8 & 9  & Llama-2-70b& 8 &9  &Tie\\
\end{tabular}
}
\end{table}

\subsection{Llama-2-70b vs. GPT 3.5}

\begin{table}[h!]
\centering
{\scriptsize
\begin{tabular}{c|ccc|ccc|c}
&\multicolumn{3}{c}{Home}&\multicolumn{3}{c}{Away} & \\[0.2cm]
\midrule       
&  & Side 1& Side 2&  &Side 1 &Side 2   &\\ 
Topic  & Winner &Llama-2-70b& GPT 3.5& Winner&GPT 3.5& Llama-2-70b &Overall\\ \midrule 
1  &GPT 3.5& 7& 9& Llama-2-70b & 8 &9&Tie\\
2  &Llama-2-70b& 8& 7& GPT 3.5& 8 &7&Tie\\
3  &Llama-2-70b& 9& 8& Llama-2-70b & 7 &8&Llama-2-70b \\
4  &GPT 3.5& 7& 8& Llama-2-70b & 7 &8&Tie\\
5  &GPT 3.5& 8& 9& GPT 3.5& 9 &8&GPT 3.5\\
6  &Llama-2-70b& 9& 8& GPT 3.5& 8 &7&Tie\\
7  &Llama-2-70b& 9& 8& GPT 3.5& 8 &7&Tie\\
8  &Llama-2-70b& 8& 7& GPT 3.5& 8 &7&Tie\\
9  &GPT 3.5& 8& 9& Llama-2-70b & 7 &8&Tie\\
10  &GPT 3.5& 7& 8& GPT 3.5& 8 &7&GPT 3.5\\
11  &GPT 3.5& 7& 8& Llama-2-70b & 7 &8&Tie\\
12  &Llama-2-70b& 8& 7& GPT 3.5& 8 &6&Tie\\
13  &GPT 3.5& 7& 8& Llama-2-70b & 7 &8&Tie\\
14  &Llama-2-70b& 9& 8& GPT 3.5& 9 &8&Tie\\
15  &GPT 3.5& 7& 8& Llama-2-70b & 7 &8&Tie\\
16  &Llama-2-70b& 9& 8& GPT 3.5& 8 &7&Tie\\
17  &Llama-2-70b& 9& 8& GPT 3.5& 8 &6&Tie\\
18  &GPT 3.5& 7& 8& Llama-2-70b & 7 &8&Tie\\
19  &GPT 3.5& 7& 8& Llama-2-70b & 7 &8&Tie\\
20  &Llama-2-70b& 8& 6& GPT 3.5& 8 &6&Tie\\
21  &Llama-2-70b& 9& 8& GPT 3.5& 8 &7&Tie\\
22  &Llama-2-70b& 8& 7& GPT 3.5& 8 &6&Tie\\
23  &Llama-2-70b& 9& 8& GPT 3.5& 8 &7&Tie\\
24  &Llama-2-70b& 9& 8& GPT 3.5& 8 &7&Tie\\
25  &Llama-2-70b& 9& 8& GPT 3.5& 9 &8&Tie\\
\end{tabular}
}
\end{table}
\newpage

\subsection{Llama-2-70b vs. GPT 4}

\begin{table}[h!]
\centering
{\scriptsize
\begin{tabular}{c|ccc|ccc|c}
&\multicolumn{3}{c}{Home}&\multicolumn{3}{c}{Away} & \\[0.2cm]
\midrule       
&  & Side 1& Side 2&  &Side 1 &Side 2   &\\ 
Topic  & Winner &Llama-2-70b& GPT 4& Winner&GPT 4& Llama-2-70b & Over\\ \midrule 
1  &Llama-2-70b& 8& 7& GPT 4& 9&7 &Tie\\
2  &GPT 4& 8& 9& Llama-2-70b& 8&9 &Tie\\
3  &GPT 4& 7& 8& GPT 4& 9&8 &GPT 4\\
4  &GPT 4& 7& 8& GPT 4& 8&7 &GPT 4\\
5  &GPT 4& 8& 9& GPT 4& 8&7 &GPT 4\\
6  &GPT 4& 8& 9& GPT 4& 8&7 &GPT 4\\
7  &GPT 4& 7& 8& GPT 4& 8&7 &GPT 4\\
8  &GPT 4& 7& 8& GPT 4& 8&7 &GPT 4\\
9  &GPT 4& 8& 9& GPT 4& 9&7 &GPT 4\\
10  &GPT 4& 7& 8& GPT 4& 8&7 &GPT 4\\
11  &GPT 4& 7& 8& GPT 4& 8&7 &GPT 4\\
12  &GPT 4& 7& 9& Llama-2-70b& 7&9 &Tie\\
13  &GPT 4& 7& 8& GPT 4& 9&7 &GPT 4\\
14  &GPT 4& 8& 9& GPT 4& 9&7 &GPT 4\\
15  &GPT 4& 8& 9& GPT 4& 8&7 &GPT 4\\
16  &GPT 4& 8& 9& Llama-2-70b& 8&9 &Tie\\
17  &GPT 4& 7& 8& tie& 8&8 &GPT 4\\
18  &GPT 4& 8& 9& GPT 4& 8&7 &GPT 4\\
19  &GPT 4& 8& 9& GPT 4& 8&7 &GPT 4\\
20  &GPT 4& 7& 9& GPT 4& 8&7 &GPT 4\\
21  &GPT 4& 7& 9& GPT 4& 8&7 &GPT 4\\
22  &GPT 4& 6& 9& GPT 4& 8&7 &GPT 4\\
23  &GPT 4& 8& 9& GPT 4& 8&7 &GPT 4\\
24  &GPT 4& 7& 9& GPT 4& 8&7 &GPT 4\\
25  &GPT 4& 8& 9& GPT 4& 9&8 &GPT 4\\
\end{tabular}
}
\end{table}

\subsection{Llama-3-70b vs. Vicuna-7b-v1.5}
\begin{table}[h!]
\centering
{\scriptsize
\begin{tabular}{c|ccc|ccc|c}
&\multicolumn{3}{c}{Home}&\multicolumn{3}{c}{Away} & \\[0.2cm]
\midrule       
&  & Side 1& Side 2&  &Side 1 &Side 2   &\\ 
Topic  & Winner &Llama-3-70b& Vicuna-7b-v1.5& Winner&Vicuna-7b-v1.5& Llama-3-70b&Overall\\ \midrule 
1  &Llama-3-70b& 8& 7& Llama-3-70b& 4&7&Llama-3-70b\\
2  &Llama-3-70b& 8& 7& Llama-3-70b& 7&9&Llama-3-70b\\
3  &Llama-3-70b& 8& 7& Llama-3-70b& 7&8&Llama-3-70b\\
4  &Llama-3-70b& 8& 7& Llama-3-70b& 7&8&Llama-3-70b\\
5  &Llama-3-70b& 9& 8& Llama-3-70b& 8&9&Llama-3-70b\\
6  &Llama-3-70b& 8& 7& Llama-3-70b& 6&8&Llama-3-70b\\
7  &Llama-3-70b& 8& 7& Llama-3-70b& 7&8&Llama-3-70b\\
8  &Llama-3-70b& 8& 7& Llama-3-70b& 6&8&Llama-3-70b\\
9  &Llama-3-70b& 9& 7& Llama-3-70b& 8&9&Llama-3-70b\\
10  &Llama-3-70b& 8& 7& Llama-3-70b& 7&8&Llama-3-70b\\
11  &Llama-3-70b& 8& 7& Llama-3-70b& 7&8&Llama-3-70b\\
12  &Llama-3-70b& 9& 7& Llama-3-70b& 6&9&Llama-3-70b\\
13  &Llama-3-70b& 9& 7& Llama-3-70b& 7&8&Llama-3-70b\\
14  &Llama-3-70b& 8& 7& Llama-3-70b& 7&8&Llama-3-70b\\
15  &Llama-3-70b& 8& 6& Llama-3-70b& 7&8&Llama-3-70b\\
16  &Llama-3-70b& 8& 7& Llama-3-70b& 7&8&Llama-3-70b\\
17  &Vicuna-7b-v1.5& 7& 8& Llama-3-70b& 6&9&Tie\\
18  &Llama-3-70b& 9& 7& Llama-3-70b& 7&8&Llama-3-70b\\
19  &Llama-3-70b& 8& 7& Vicuna-7b-v1.5& 8&7&Tie\\
20  &Llama-3-70b& 8& 7& Llama-3-70b& 6&9&Llama-3-70b\\
21  &Llama-3-70b& 8& 7& Llama-3-70b& 6&8&Llama-3-70b\\
22  &Llama-3-70b& 8& 7& Llama-3-70b& 6&9&Llama-3-70b\\
23  &Llama-3-70b& 8& 6& Llama-3-70b& 6&8&Llama-3-70b\\
24  &Llama-3-70b& 8& 7& Llama-3-70b& 7&9&Llama-3-70b\\
25  &Llama-3-70b& 8& 7& Llama-3-70b& 7&9&Llama-3-70b\\
\end{tabular}
}
\end{table}
\newpage

\subsection{Llama-3-70b vs. Vicuna-13b-v1.5}
\begin{table}[h!]
\centering
{\scriptsize
\begin{tabular}{c|ccc|ccc|c}
&\multicolumn{3}{c}{Home}&\multicolumn{3}{c}{Away} & \\[0.2cm]
\midrule       
&  & Side 1& Side 2&  &Side 1 &Side 2   &\\ 
Topic  & Winner &Llama-3-70b& Vicuna-13b-v1.5& Winner&Vicuna-13b-v1.5& Llama-3-70b&Overall\\ \midrule 
1  &Llama-3-70b& 9& 7& Llama-3-70b& 8&9&Llama-3-70b\\
2  &Vicuna-13b-v1.5& 7& 8& Llama-3-70b& 7&8&Tie\\
3  &Vicuna-13b-v1.5& 8& 9& Llama-3-70b& 7&8&Tie\\
4  &Llama-3-70b& 8& 7& Llama-3-70b& 7&8&Llama-3-70b\\
5  &Llama-3-70b& 8& 7& Llama-3-70b& 8&9&Llama-3-70b\\
6  &Vicuna-13b-v1.5& 8& 9& Llama-3-70b& 7&8&Tie\\
7  &Llama-3-70b& 8& 7& Llama-3-70b& 7&9&Llama-3-70b\\
8  &Llama-3-70b& 8& 7& Llama-3-70b& 6&8&Llama-3-70b\\
9  &Llama-3-70b& 8& 7& Llama-3-70b& 7&8&Llama-3-70b\\
10  &tie& 8& 8& Llama-3-70b& 7&8&Tie\\
11  &Llama-3-70b& 8& 7& Llama-3-70b& 8&9&Llama-3-70b\\
12  &Vicuna-13b-v1.5& 8& 9& Llama-3-70b& 6&9&Tie\\
13  &Llama-3-70b& 8& 7& Llama-3-70b& 7&9&Llama-3-70b\\
14  &Llama-3-70b& 9& 8& Llama-3-70b& 7&8&Llama-3-70b\\
15  &Llama-3-70b& 9& 7& Llama-3-70b& 7&8&Llama-3-70b\\
16  &tie& 8& 8& Llama-3-70b& 7&9&Llama-3-70b\\
17  &Vicuna-13b-v1.5& 7& 8& Llama-3-70b& 6&9&Tie\\
18  &Llama-3-70b& 9& 7& Llama-3-70b& 7&8&Llama-3-70b\\
19  &Llama-3-70b& 9& 7& Llama-3-70b& 8&9&Llama-3-70b\\
20  &Vicuna-13b-v1.5& 7& 8& Llama-3-70b& 6&9&Tie\\
21  &tie& 8& 8& Llama-3-70b& 7&9&Llama-3-70b\\
22  &Vicuna-13b-v1.5& 7& 8& Llama-3-70b& 6&8&Tie\\
23  &Llama-3-70b& 8& 7& Llama-3-70b& 7&8&Llama-3-70b\\
24  &Llama-3-70b& 9& 8& Llama-3-70b& 7&8&Llama-3-70b\\
25  &Llama-3-70b& 8& 7& Llama-3-70b& 8&9&Llama-3-70b\\
\end{tabular}
}
\end{table}

\subsection{Llama-3-70b vs. Mixtral-8x7B}

\begin{table}[h!]
\centering
{\scriptsize
\begin{tabular}{c|ccc|ccc|c}
&\multicolumn{3}{c}{Home}&\multicolumn{3}{c}{Away} & \\[0.2cm]
\midrule       
&  & Side 1& Side 2&  &Side 1 &Side 2   &\\ 
Topic  & Winner &Llama-3-70b& Mixtral-8x7B& Winner&Mixtral-8x7B& Llama-3-70b&Overall\\ \midrule 
1  &Llama-3-70b& 9& 8& Mixtral-8x7B& 8&7&Tie\\
2  &Llama-3-70b& 8& 7& Llama-3-70b& 8&9&Llama-3-70b\\
3  &Llama-3-70b& 8& 7& Llama-3-70b& 8&9&Llama-3-70b\\
4  &Llama-3-70b& 8& 7& Mixtral-8x7B& 8&7&Tie\\
5  &Llama-3-70b& 8& 7& Llama-3-70b& 7&8&Llama-3-70b\\
6  &Mixtral-8x7B& 8& 9& Llama-3-70b& 8&9&Tie\\
7  &Mixtral-8x7B& 7& 8& Llama-3-70b& 7&8&Tie\\
8  &Llama-3-70b& 8& 7& Llama-3-70b& 7&8&Llama-3-70b\\
9  &Mixtral-8x7B& 8& 9& Llama-3-70b& 8&9&Tie\\
10  &Llama-3-70b& 8& 7& Llama-3-70b& 8&9&Llama-3-70b\\
11  &Llama-3-70b& 8& 7& Mixtral-8x7B& 8&7&Tie\\
12  &Mixtral-8x7B& 7& 9& Llama-3-70b& 7&8&Tie\\
13  &Llama-3-70b& 8& 7& Llama-3-70b& 7&8&Llama-3-70b\\
14  &Llama-3-70b& 8& 7& Mixtral-8x7B& 8&7&Tie\\
15  &Llama-3-70b& 9& 8& Mixtral-8x7B& 8&7&Tie\\
16  &Mixtral-8x7B& 8& 9& Llama-3-70b& 8&9&Tie\\
17  &Mixtral-8x7B& 8& 9& Llama-3-70b& 7&8&Tie\\
18  &Llama-3-70b& 9& 8& Mixtral-8x7B& 8&7&Tie\\
19  &Llama-3-70b& 8& 7& Mixtral-8x7B& 8&7&Tie\\
20  &Mixtral-8x7B& 7& 8& Llama-3-70b& 6&8&Tie\\
21  &Mixtral-8x7B& 7& 8& Llama-3-70b& 8&9&Tie\\
22  &Mixtral-8x7B& 7& 8& Llama-3-70b& 6&9&Tie\\
23  &Mixtral-8x7B& 7& 8& Llama-3-70b& 8&9&Tie\\
24  &Mixtral-8x7B& 8& 9& Llama-3-70b& 7&8&Tie\\
25  &Mixtral-8x7B& 7& 8& Llama-3-70b& 7&8&Tie\\
\end{tabular}
}
\end{table}
\newpage

\subsection{Llama-3-70b vs. GPT-3.5}
\begin{table}[h!]
\centering
{\scriptsize
\begin{tabular}{c|ccc|ccc|c}
&\multicolumn{3}{c}{Home}&\multicolumn{3}{c}{Away} & \\[0.2cm]
\midrule       
&  & Side 1& Side 2&  &Side 1 &Side 2   &\\ 
Topic  & Winner &Llama-3-70b& GPT-3.5& Winner&GPT-3.5& Llama-3-70b&Overall\\ \midrule 
1  &Llama-3-70b& 8& 7& GPT-3.5& 8&7&Tie\\
2  &GPT-3.5& 8& 9& Llama-3-70b& 7&8&Tie\\
3  &GPT-3.5& 8& 9& Llama-3-70b& 8&9&Tie\\
4  &Llama-3-70b& 8& 7& GPT-3.5& 8&7&Tie\\
5  &tie& 8& 8& Llama-3-70b& 8&9&Llama-3-70b\\
6  &GPT-3.5& 7& 8& Llama-3-70b& 7&8&Tie\\
7  &GPT-3.5& 8& 9& Llama-3-70b& 7&9&Tie\\
8  &GPT-3.5& 7& 8& Llama-3-70b& 6&8&Tie\\
9  &Llama-3-70b& 8& 7& GPT-3.5& 8&7&Tie\\
10  &GPT-3.5& 8& 9& Llama-3-70b& 8&9&Tie\\
11  &Llama-3-70b& 8& 7& GPT-3.5& 8&7&Tie\\
12  &GPT-3.5& 8& 9& Llama-3-70b& 6&9&Tie\\
13  &Llama-3-70b& 8& 7& GPT-3.5& 8&7&Tie\\
14  &Llama-3-70b& 8& 7& Llama-3-70b& 7&8&Llama-3-70b\\
15  &Llama-3-70b& 9& 8& GPT-3.5& 8&7&Tie\\
16  &Llama-3-70b& 9& 8& Llama-3-70b& 7&8&Llama-3-70b\\
17  &GPT-3.5& 8& 9& Llama-3-70b& 7&9&Tie\\
18  &Llama-3-70b& 8& 7& Llama-3-70b& 8&9&Llama-3-70b\\
19  &Llama-3-70b& 8& 7& GPT-3.5& 8&7&Tie\\
20  &GPT-3.5& 7& 8& Llama-3-70b& 7&9&Tie\\
21  &GPT-3.5& 8& 9& Llama-3-70b& 7&8&Tie\\
22  &GPT-3.5& 7& 9& Llama-3-70b& 6&9&Tie\\
23  &Llama-3-70b& 8& 7& Llama-3-70b& 7&8&Llama-3-70b\\
24  &GPT-3.5& 8& 9& Llama-3-70b& 7&8&Tie\\
25  &GPT-3.5& 8& 9& Llama-3-70b& 8&9&Tie\\
\end{tabular}
}
\end{table}

\subsection{Llama-3-70b vs. GPT-4}

\begin{table}[h!]
\centering
{\scriptsize
\begin{tabular}{c|ccc|ccc|c}
&\multicolumn{3}{c}{Home}&\multicolumn{3}{c}{Away} & \\[0.2cm]
\midrule       
&  & Side 1& Side 2&  &Side 1 &Side 2   &\\ 
Topic  & Winner &Llama-3-70b& GPT-4& Winner&GPT-4& Llama-3-70b&Overall\\ \midrule 
1  &Llama-3-70b& 8& 7& GPT-4& 9&7&Tie\\
2  &GPT-4& 7& 9& Llama-3-70b& 7&8&Tie\\
3  &GPT-4& 8& 9& GPT-4& 8&7&GPT-4\\
4  &GPT-4& 7& 8& GPT-4& 8&7&GPT-4\\
5  &GPT-4& 8& 9& GPT-4& 8&7&GPT-4\\
6  &GPT-4& 7& 8& Llama-3-70b& 8&9&Tie\\
7  &GPT-4& 8& 9& GPT-4& 8&7&GPT-4\\
8  &GPT-4& 7& 8& GPT-4& 8&7&GPT-4\\
9  &GPT-4& 8& 9& GPT-4& 8&7&GPT-4\\
10  &GPT-4& 8& 9& Llama-3-70b& 8&9&Tie\\
11  &Llama-3-70b& 8& 7& GPT-4& 8&7&Tie\\
12  &GPT-4& 7& 8& Llama-3-70b& 8&9&Tie\\
13  &Llama-3-70b& 8& 7& GPT-4& 8&7&Tie\\
14  &GPT-4& 8& 9& GPT-4& 9&8&GPT-4\\
15  &GPT-4& 8& 9& GPT-4& 9&7&GPT-4\\
16  &GPT-4& 8& 9& Llama-3-70b& 8&9&Tie\\
17  &GPT-4& 8& 9& GPT-4& 8&7&GPT-4\\
18  &GPT-4& 8& 9& GPT-4& 8&7&GPT-4\\
19  &Llama-3-70b& 8& 7& GPT-4& 9&8&Tie\\
20  &GPT-4& 7& 9& Llama-3-70b& 7&8&Tie\\
21  &GPT-4& 7& 8& Llama-3-70b& 8&9&Tie\\
22  &GPT-4& 7& 9& GPT-4& 8&7&GPT-4\\
23  &GPT-4& 7& 9& GPT-4& 8&7&GPT-4\\
24  &GPT-4& 8& 9& Llama-3-70b& 8&9&Tie\\
25  &GPT-4& 8& 9& GPT-4& 8&7&GPT-4\\
\end{tabular}
}
\end{table}

\newpage





\subsection{Vicuna-7b-v1.5 vs. Vicuna-7b-v1.5}

\begin{table}[h!]
\centering
{\scriptsize
\begin{tabular}{c|ccc|ccc}
&\multicolumn{3}{c}{Home}&\multicolumn{3}{c}{Away} \\[0.2cm]
\midrule       
&  & Side 1& Side 2&  &Side 1 &Side 2  \\ 
Topic  & Winner &Vicuna-7b-v1.5& Vicuna-7b-v1.5& Winner&Vicuna-7b-v1.5& Vicuna-7b-v1.5\\ \midrule 
1  &1& 8& 7& 1& 8&7\\
2  &2& 7& 8& 2& 7&8\\
3  &1& 8& 7& 1& 8&7\\
4  &1& 8& 7& 1& 8&7\\
5  &1& 8& 7& 1& 8&7\\
6  &1& 8& 7& 2& 7&8\\
7  &1& 8& 7& 2& 8&9\\
8  &2& 7& 8& 2& 7&8\\
9  &1& 8& 7& 1& 8&7\\
10  &1& 8& 7& 1& 8&7\\
11  &1& 8& 7& 1& 8&7\\
12  &2& 6& 8& 2& 7&8\\
13  &2& 7& 8& 2& 7&8\\
14  &2& 7& 8& 1& 8&7\\
15  &1& 8& 7& 1& 8&7\\
16  &1& 8& 7& 1& 8&7\\
17  &2& 8& 9& 2& 7&8\\
18  &1& 8& 7& 1& 8&7\\
19  &1& 8& 7& 1& 8&7\\
20  &2& 7& 8& 2& 6&8\\
21  &1& 8& 7& 1& 8&7\\
22  &2& 7& 8& 2& 6&8\\
23  &1& 8& 7& 1& 8&7\\
24  &1& 8& 7& 1& 8&7\\
25  &2& 8& 9& 2& 8&9\\
\end{tabular}
}
\end{table}

\subsection{Vicuna-7b-v1.5 vs. Vicuna-13b-v1.5}

\begin{table}[h!]
\centering
{\scriptsize
\begin{tabular}{c|ccc|cccl}
&\multicolumn{3}{c}{Home}&\multicolumn{3}{c}{Away} & \\[0.2cm]
\midrule       
&  & Side 1& Side 2&  &Side 1 &Side 2   &\\ 
Topic  & Winner &Vicuna-7b-v1.5& Vicuna-13b-v1.5& Winner&Vicuna-13b-v1.5& Vicuna-7b-v1.5 &Overall\\ \midrule 
1  &Vicuna-7b-v1.5& 8& 7& Vicuna-13b-v1.5& 8&7 &Tie\\
2  &Vicuna-13b-v1.5& 7& 8& Vicuna-7b-v1.5& 7&8 &Tie\\
3  &Vicuna-13b-v1.5& 8& 9& Vicuna-13b-v1.5& 8&7 &Vicuna-13b-v1.5\\
4  &Vicuna-7b-v1.5& 8& 7& Vicuna-13b-v1.5& 8&7 &Tie\\
5  &Vicuna-7b-v1.5& 8& 7& Vicuna-13b-v1.5& 8&7 &Tie\\
6  &Vicuna-13b-v1.5& 6& 8& Vicuna-7b-v1.5& 7&8 &Tie\\
7  &Vicuna-13b-v1.5& 8& 9& Vicuna-7b-v1.5& 8&9 &Tie\\
8  &Vicuna-13b-v1.5& 7& 8& Vicuna-7b-v1.5& 7&8 &Tie\\
9  &Vicuna-7b-v1.5& 8& 7& Vicuna-13b-v1.5& 8&7 &Tie\\
10  &Vicuna-13b-v1.5& 8& 9& tie& 8&8 &Vicuna-13b-v1.5\\
11  &Vicuna-7b-v1.5& 8& 7& Vicuna-13b-v1.5& 8&7 &Tie\\
12  &Vicuna-13b-v1.5& 6& 8& Vicuna-7b-v1.5& 6&8 &Tie\\
13  &Vicuna-7b-v1.5& 8& 7& Vicuna-13b-v1.5& 8&7 &Tie\\
14  &Vicuna-13b-v1.5& 8& 9& Vicuna-13b-v1.5& 8&7 &Vicuna-13b-v1.5\\
15  &Vicuna-7b-v1.5& 8& 7& Vicuna-13b-v1.5& 8&7 &Tie\\
16  &Vicuna-13b-v1.5& 7& 8& Vicuna-7b-v1.5& 8&9 &Tie\\
17  &Vicuna-13b-v1.5& 7& 8& Vicuna-7b-v1.5& 7&8 &Tie\\
18  &Vicuna-7b-v1.5& 8& 7& Vicuna-13b-v1.5& 8&6 &Tie\\
19  &Vicuna-7b-v1.5& 8& 7& Vicuna-13b-v1.5& 8&7 &Tie\\
20  &Vicuna-13b-v1.5& 7& 8& Vicuna-7b-v1.5& 6&7 &Tie\\
21  &Vicuna-13b-v1.5& 7& 8& Vicuna-7b-v1.5& 7&8 &Tie\\
22  &Vicuna-13b-v1.5& 6& 8& Vicuna-7b-v1.5& 7&8 &Tie\\
23  &Vicuna-13b-v1.5& 7& 8& Vicuna-7b-v1.5& 7&8 &Tie\\
24  &Vicuna-7b-v1.5& 8& 7& Vicuna-13b-v1.5& 8&7 &Tie\\
25  &Vicuna-7b-v1.5& 8& 7& Vicuna-13b-v1.5& 8&7 &Tie\\
\end{tabular}
}
\end{table}
\newpage

\subsection{Vicuna-7b-v1.5 vs. Mixtral-8x7B}

\begin{table}[h!]
\centering
{\scriptsize
\begin{tabular}{c|ccc|ccc|c}
&\multicolumn{3}{c}{Home}&\multicolumn{3}{c}{Away} & \\[0.2cm]
\midrule       
&  & Side 1& Side 2&  &Side 1 &Side 2   &\\ 
Topic  & Winner &Vicuna-7b-v1.5& Mixtral-8x7B& Winner&Mixtral-8x7B& Vicuna-7b-v1.5 &Overall\\ \midrule 
1  &Vicuna-7b-v1.5& 8& 7& Mixtral-8x7B& 9&7 &Tie\\
2  &Mixtral-8x7B& 7& 8& Mixtral-8x7B& 8&7 &Mixtral-8x7B\\
3  &Vicuna-7b-v1.5& 8& 7& Mixtral-8x7B& 8&7 &Tie\\
4  &Vicuna-7b-v1.5& 8& 7& Mixtral-8x7B& 8&7 &Tie\\
5  &tie& 8& 8& Mixtral-8x7B& 9&7 &Mixtral-8x7B\\
6  &Mixtral-8x7B& 7& 8& Mixtral-8x7B& 8&7 &Mixtral-8x7B\\
7  &Mixtral-8x7B& 7& 8& Vicuna-7b-v1.5& 8&9 &Tie\\
8  &Mixtral-8x7B& 7& 8& Mixtral-8x7B& 8&7 &Mixtral-8x7B\\
9  &Vicuna-7b-v1.5& 8& 7& Mixtral-8x7B& 9&7 &Tie\\
10  &Mixtral-8x7B& 7& 8& Mixtral-8x7B& 8&7 &Mixtral-8x7B\\
11  &Vicuna-7b-v1.5& 8& 7& Mixtral-8x7B& 8&6 &Tie\\
12  &Mixtral-8x7B& 6& 8& Mixtral-8x7B& 8&7 &Mixtral-8x7B\\
13  &Vicuna-7b-v1.5& 8& 7& Mixtral-8x7B& 8&6 &Tie\\
14  &Mixtral-8x7B& 8& 9& Mixtral-8x7B& 8&7 &Mixtral-8x7B\\
15  &Vicuna-7b-v1.5& 8& 7& Mixtral-8x7B& 8&7 &Tie\\
16  &Mixtral-8x7B& 7& 8& Vicuna-7b-v1.5& 8&9 &Tie\\
17  &Mixtral-8x7B& 7& 9& Vicuna-7b-v1.5& 8&9 &Tie\\
18  &Vicuna-7b-v1.5& 8& 7& Mixtral-8x7B& 8&7 &Tie\\
19  &Vicuna-7b-v1.5& 8& 7& Mixtral-8x7B& 8&7 &Tie\\
20  &Mixtral-8x7B& 7& 9& Vicuna-7b-v1.5& 7&8 &Tie\\
21  &Mixtral-8x7B& 7& 8& Mixtral-8x7B& 8&7 &Mixtral-8x7B\\
22  &Mixtral-8x7B& 6& 8& Vicuna-7b-v1.5& 7&8 &Tie\\
23  &Vicuna-7b-v1.5& 8& 7& Mixtral-8x7B& 8&7 &Tie\\
24  &Mixtral-8x7B& 7& 8& Mixtral-8x7B& 8&7 &Mixtral-8x7B\\
25  &Mixtral-8x7B& 7& 8& Mixtral-8x7B& 8&8 &Mixtral-8x7B\\
\end{tabular}
}
\end{table}

\subsection{Vicuna-7b-v1.5 vs. GPT 3.5}

\begin{table}[h!]
\centering
{\scriptsize
\begin{tabular}{c|ccc|ccc|c}
&\multicolumn{3}{c}{Home}&\multicolumn{3}{c}{Away} & \\[0.2cm]
\midrule       
&  & Side 1& Side 2&  &Side 1 &Side 2   &\\ 
Topic  & Winner &Vicuna-7b-v1.5& GPT 3.5& Winner&GPT 3.5& Vicuna-7b-v1.5 &Overall\\ \midrule 
1  &Vicuna-7b-v1.5& 8& 7& GPT 3.5& 9&8 &Tie\\
2  &GPT 3.5& 7& 8& Vicuna-7b-v1.5 & 7&8 &Tie\\
3  &GPT 3.5& 8& 7& GPT 3.5& 8&7 &GPT 3.5\\
4  &GPT 3.5& 8& 7& GPT 3.5& 8&7 &GPT 3.5\\
5  &GPT 3.5& 8& 8& GPT 3.5& 8&7 &GPT 3.5\\
6  &GPT 3.5& 7& 8& tie& 8&8 &GPT 3.5\\
7  &GPT 3.5& 7& 8& GPT 3.5& 8&7 &GPT 3.5\\
8  &GPT 3.5& 7& 8& GPT 3.5& 8&7 &GPT 3.5\\
9  &GPT 3.5& 8& 7& GPT 3.5& 8&7 &GPT 3.5\\
10  &GPT 3.5& 7& 8& GPT 3.5& 8&7 &GPT 3.5\\
11  &GPT 3.5& 8& 7& GPT 3.5& 8&7 &GPT 3.5\\
12  &GPT 3.5& 6& 8& Vicuna-7b-v1.5 & 7&8 &Tie\\
13  &GPT 3.5& 8& 7& GPT 3.5& 8&7 &GPT 3.5\\
14  &GPT 3.5& 8& 9& GPT 3.5& 8&7 &GPT 3.5\\
15  &Vicuna-7b-v1.5& 8& 7& GPT 3.5& 9&6 &Tie\\
16  &GPT 3.5& 7& 8& GPT 3.5& 8&7 &GPT 3.5\\
17  &GPT 3.5& 7& 9& Vicuna-7b-v1.5 & 7&8 &Tie\\
18  &Vicuna-7b-v1.5& 8& 7& GPT 3.5& 8&7 &Tie\\
19  &Vicuna-7b-v1.5& 8& 7& GPT 3.5& 8&7 &Tie\\
20  &GPT 3.5& 7& 9& GPT 3.5& 8&7 &GPT 3.5\\
21  &GPT 3.5& 7& 8& GPT 3.5& 8&7 &GPT 3.5\\
22  &GPT 3.5& 6& 8& GPT 3.5& 8&7 &GPT 3.5\\
23  &GPT 3.5& 8& 7& GPT 3.5& 8&7 &GPT 3.5\\
24  &GPT 3.5& 7& 8& GPT 3.5& 8&7 &GPT 3.5\\
25  &GPT 3.5& 7& 8& tie& 8&8 &GPT 3.5\\
\end{tabular}
}
\end{table}

\newpage

\subsection{Vicuna-7b-v1.5 vs. GPT 4}

\begin{table}[h!]
\centering
{\scriptsize
%
}
\end{table}

\subsection{Vicuna-13b-v1.5 vs. Vicuna-13b-v1.5}

\begin{table}[h!]
\centering
{\scriptsize
%
}
\end{table}
\newpage

\subsection{Vicuna-13b-v1.5 vs. Mixtral-8x7B}

\begin{table}[h!]
\centering
{\scriptsize
%
}
\end{table}

\subsection{Vicuna-13b-v1.5 vs. GPT 3.5}

\begin{table}[h!]
\centering
{\scriptsize
%
}
\end{table}
\newpage

\subsection{Vicuna-13b-v1.5 vs. GPT 4}

\begin{table}[h!]
\centering
{\scriptsize
%
}
\end{table}

\subsection{Mixtral-8x7B vs. Mixtral-8x7B}
~
\begin{table}[h!]
\centering
{\scriptsize
%
}
\end{table}

\newpage

\subsection{Mixtral-8x7B vs. GPT 3.5}

\begin{table}[h!]
\centering
{\scriptsize
%
}
\end{table}

\subsection{Mixtral-8x7B vs. GPT 4}

\begin{table}[h!]
\centering
{\scriptsize
%
}
\end{table}
\newpage

\subsection{GPT 3.5 vs. GPT 3.5}

\begin{table}[h!]
\centering
{\scriptsize
%
}
\end{table}

\subsection{GPT 3.5 vs. GPT 4}

\begin{table}[h!]
\centering
{\scriptsize
%
}
\end{table}
\newpage

\subsection{GPT 4 vs. GPT 4}

\begin{table}[h!]
\centering
{\scriptsize
%
}
\end{table}

\section{Other Judges}
In this section, we report the results of experiment from Section~\ref{sec:other} with Llama-3-70b as judge.

\subsection{Llama2-13b vs. Llama2-7b}

\begin{table}[h!]
\centering
{\scriptsize
%
}
\end{table}

\newpage
\subsection{Llama2-13b vs. Llama2-70b}

\begin{table}[h!]
\centering
{\scriptsize
%
}
\end{table}

\subsection{Llama2-13b vs. Llama3-70b}

\begin{table}[h!]
\centering
{\scriptsize
%
}
\end{table}
\newpage

\subsection{Llama2-13b vs. Vicuna-7b-v1.5}

\begin{table}[h!]
\centering
{\scriptsize
%
}
\end{table}
\subsection{Llama2-13b vs. Vicuna-13b-v1.5}

\begin{table}[h!]
\centering
{\scriptsize
%
}
\end{table}
\newpage

\subsection{Llama2-13b vs. Mixtral-8x7B}

\begin{table}[h!]
\centering
{\scriptsize
%
}
\end{table}

\subsection{Llama2-13b vs. GPT-3.5}

\begin{table}[h!]
\centering
{\scriptsize
%
}
\end{table}
\newpage

\subsection{Llama2-13b vs. GPT-4}

\begin{table}[h!]
\centering
{\scriptsize
%
}
\end{table}

\end{document}